\documentclass[11pt]{article}

\usepackage[T1]{fontenc}
\usepackage[final]{microtype}
\usepackage{csquotes,lmodern}
\usepackage[colorlinks=true,linkcolor=blue,citecolor=blue,urlcolor=blue]{hyperref}
\usepackage[most]{tcolorbox}

\usepackage{multicol} 
\usepackage{multirow}

\hypersetup{hidelinks}
\usepackage{indentfirst}

\usepackage[numbers,sort&compress]{natbib}

\usepackage{authblk}

\usepackage[margin=1in]{geometry}
\usepackage{times} 
\usepackage{microtype}
\usepackage{graphicx}
\usepackage{amsmath,amssymb}
\usepackage{booktabs}
\usepackage{hypcap}
\usepackage{xcolor}
\usepackage{titlesec}

\usepackage{caption}

\usepackage{float}

\usepackage[font=small]{caption}
\definecolor{mygreen}{RGB}{0,100,0} 
\definecolor{myred}{RGB}{161, 58, 58}
\usepackage{float}
\newcommand{\subfigref}[2]{\hyperref[#1]{\ref*{#1}#2}} 

\usepackage{titlesec}

\titlespacing{\section}
  {0pt}        
  {1.8ex plus 0.2ex minus 0.1ex}   
  {0.8ex plus 0.1ex minus 0.05ex}  

\titlespacing{\subsection}
  {0pt}
  {2.0ex plus 0.4ex minus 0.2ex}  
  {1.0ex plus 0.05ex}

\titlespacing{\subsubsection}
  {0pt}
  {1.5ex plus 0.4ex minus 0.2ex}  
  {0.5ex plus 0.05ex minus 0.05ex}

\titleformat{\section}
  {\normalfont\Large\bfseries}{\thesection}{1em}{}
\titleformat{\subsection}
  {\normalfont\large\bfseries}{\thesubsection}{1em}{}
\titleformat{\subsubsection}
  {\normalfont\normalsize\bfseries}{\thesubsubsection}{1em}{}

\hypersetup{
    colorlinks=true,
    linkcolor=blue,
    citecolor=blue,
    urlcolor=blue,
    pdftitle={MIRIAD: Augmenting LLMs with millions of medical query-response pairs}
}

\title{MIRIAD: Augmenting LLMs with millions of medical query-response pairs}

\author[1]{Qinyue Zheng\thanks{These authors contributed equally to this work.}}
\author[2]{Salman Abdullah$^*$}
\author[3]{Sam Rawal}
\author[4]{Cyril Zakka}
\author[2,5]{Sophie Ostmeier}
\author[6]{Maximilian Purk}
\author[7]{Eduardo Reis}
\author[8]{Eric J. Topol}
\author[2]{Jure Leskovec}
\author[1]{Michael Moor\thanks{Corresponding author: \texttt{michael.moor@bsse.ethz.ch}}}

\affil[1]{Department of Biosystems Science and Engineering, ETH Zurich, Basel, Switzerland}
\affil[2]{Department of Computer Science, Stanford University, Stanford, CA, USA}
\affil[3]{Department of Internal Medicine, Mayo Clinic, Phoenix, AZ, USA}
\affil[4]{Hugging Face, New York City, NY, USA}
\affil[5]{Department of Radiology, Stanford University, Stanford, CA, USA}
\affil[6]{Hasso-Plattner-Institute for Digital Engineering, University of Potsdam, Potsdam, Germany}
\affil[7]{Center for Artificial Intelligence in Medicine and Imaging, Stanford, CA, USA}
\affil[8]{Scripps Translational Science Institute, San Diego, CA, USA}

\date{}

\begin{document}
\maketitle

\vspace{-1.5em}
\begin{abstract}
\noindent
Large language models (LLMs) are bound to transform healthcare with advanced decision support and flexible chat assistants. However, LLMs are prone to generate inaccurate medical content. In order to ground LLMs in high-quality medical knowledge, LLMs have been equipped with external knowledge sources via retrieval augmented generation (RAG), where unstructured medical knowledge is split into small chunks of text that can be selectively retrieved and integrated into the LLMs context. Yet, existing RAG pipelines rely on raw, unstructured medical text, which can be noisy, uncurated, and difficult for LLMs to effectively leverage. Systematic approaches to organize medical knowledge and to best surface it to LLMs are generally lacking. To address these challenges, here, we introduce MIRIAD, a large-scale, curated corpus of 5,821,948 medical instruction-response pairs, each rephrased from and grounded in a passage from peer-reviewed medical literature using a semi-automated pipeline combining LLM generation, filtering, grounding, and human annotation. Unlike prior medical corpora, which rely on unstructured text, MIRIAD encapsulates rich and web-scale medical knowledge in an operationalized query-response format, which enables more targeted retrieval. Experiments on challenging medical question-answering benchmarks show that augmenting LLMs with MIRIAD improves accuracy up to 6.7\% compared to unstructured RAG baselines with the same source corpus and with the same amount of retrieved text. Moreover, MIRIAD improved the ability of LLMs to detect medical hallucinations by 22.5 to 37\% (increase in F1 score). We further introduce MIRIAD-Atlas, an interactive semantic map of MIRIAD spanning 56 medical disciplines, enabling clinical users to visually explore, search, and refine medical knowledge. MIRIAD promises to unlock a wealth of down-stream applications, including medical information retrievers, enhanced RAG applications, and knowledge-grounded chat interfaces, which ultimately enables more reliable LLM applications in healthcare.
\end{abstract}
\vspace{1em}

\section{Introduction}
Large language models (LLMs) have demonstrated striking performance in a wide range of natural language processing tasks such as question answering, translation, or summarization\cite{brown2020language, achiam2023gpt, brants2007large, devlin2019bert, radford2019language, touvron2023llama}. However, they struggle with maintaining factual correctness and staying up-to-date with domain-specific knowledge \cite{achiam2023gpt, li2023halueval, pal2023med}. This limitation is particularly pronounced in the high-stakes domain of healthcare, where factual accuracy can become a matter of life and death. Previous works leveraged retrieval augmented generation (RAG) as an inexpensive solution to ground LLMs in relevant medical literature which can reduce the need for costly LLM fine-tuning scenarios \cite{zakka2024almanac, lievin2024can}. 

Prior retrieval solutions predominantly rely on general-domain text embeddings and off-the-shelf vector databases. Achieving high retrieval performance with RAG can be hard as search queries and the corresponding relevant documents or passages may be lexically as well as even semantically distinct from each other. Recent state-of-the-art general-domain retrieval models such as ColBERT, ColBERTv2, Jina-ColBERT-v2, or E5 have demonstrated the value of retrieval training—having been trained on large, supervised retrieval datasets, such as MS Marco and CC Pairs \cite{khattab2020colbert, wang2022text, jha2024jina, santhanam2022colbertv2, bajaj2016ms}. That is, these datasets typically comprise paired samples of queries and documents (or questions and answers). However, the medical domain still lacks high-quality, large-scale, and openly accessible retrieval datasets, i.e., datasets with paired instructions and responses—that could enable the development of retrieval systems optimized for medical information. 

Existing medical question-answer (QA) datasets like PubMedQA, MedMCQA, or MedQA fall short in several ways \cite{jin2019pubmedqa,pal2022medmcqa, jin2021disease}. For example, PubMedQA focuses on specific article sections and lacks free-form answers and MedQA just contains multiple-choice questions (MCQs). Furthermore, existing medical QA datasets are of a limited size, typically ranging from thousands to hundreds of thousands of samples \cite{jin2019pubmedqa,pal2022medmcqa, jin2021disease, kim2024medexqa, krithara2023bioasq}. 

To overcome these limitations, here, we present MIRIAD, which stands for a “Medical Instruction and Retrieval Dataset”. MIRIAD represents a large-scale resource of operationalized medical knowledge in the form of medical instructions and responses that were semi-synthetically generated using LLMs, each pair grounded in articles from the medical literature. To ensure high quality, we conducted a sequence of quality control steps, including rule-based filtering, LLM-based filtering and human expert annotations. MIRIAD is released as a series of two versions: MIRIAD-5.8M (5,821,948 samples after rule-based filtering) and MIRIAD-4.4M (4,487,542 samples after the full sequence of quality control steps). Due to the employed literature rephrasing approach, resulting QA pairs are grounded in the peer-reviewed medical literature. In our experiments, we find that MIRIAD can directly be utilized as an external corpus of knowledge to improve medical retrieval augmented generation (RAG) performance in LLMs compared to the unstructured text of the same source by up to 6.7\%. MIRIAD can be directly used for training medical information retrieval models to further improve retrieval quality. MIRIAD improved the ability of LLMs to detect medical hallucinations by 22.5 to 37\% (measured with F1 score). By releasing MIRIAD-Atlas, an interactive web-hosted landscape of operationalized medical knowledge, we make MIRIAD broadly accessible as a resource for users to visually explore, search, and refine medical knowledge surfaced as a landscape containing millions of medical queries and responses, organized by topic and disciplines. Finally, we expect that MIRIAD will enable the development of advanced medical retrieval systems, enhanced RAG applications, knowledge-grounded clinical AI chat interfaces, and ultimately empower researchers, caregivers, as well as patients.

\section{Results}
\subsection{The MIRIAD dataset}

\subsubsection*{Dataset creation} 

\begin{figure}
    \centering
    \includegraphics[width=1\linewidth]{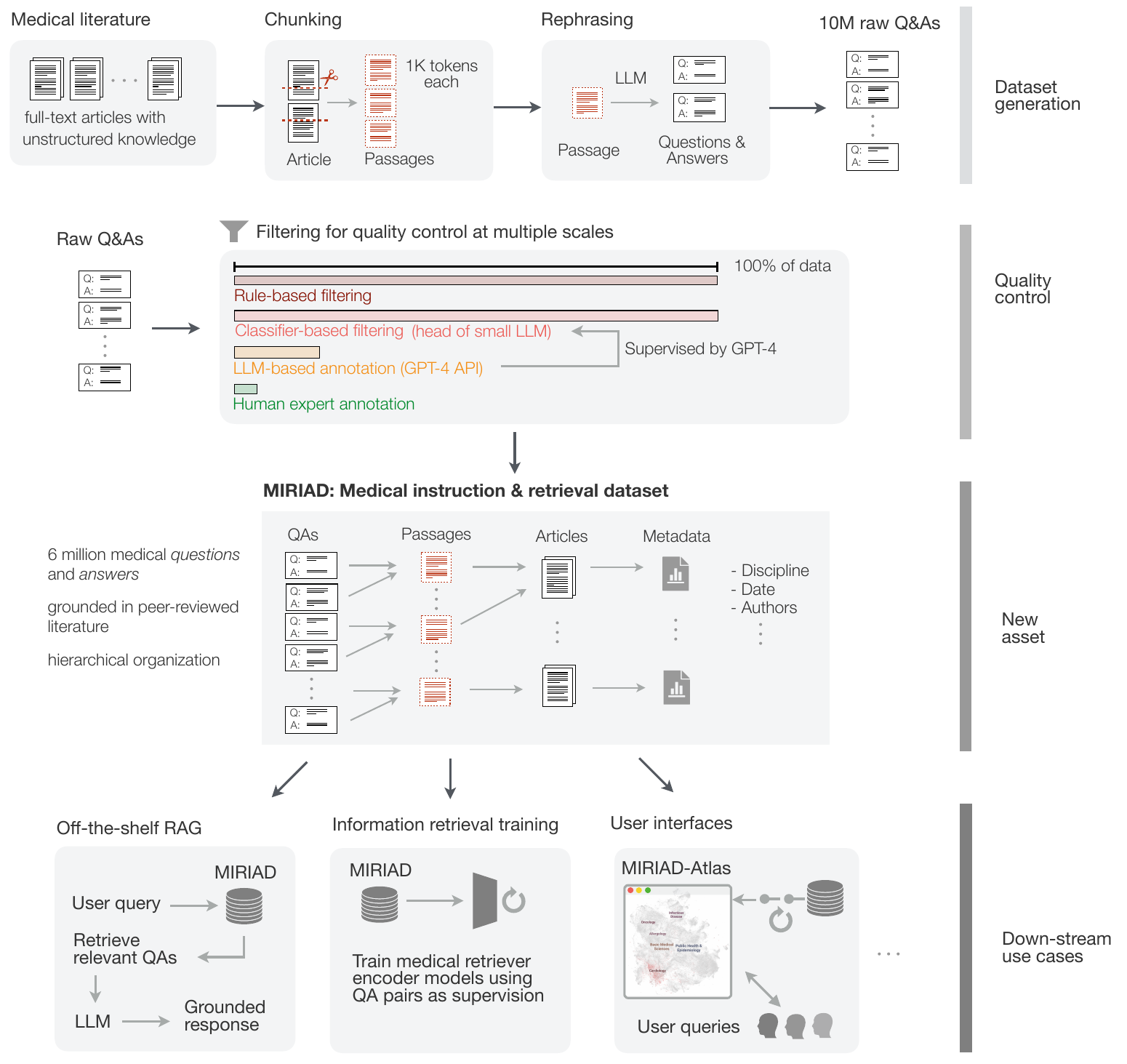}
    \caption{\textbf{Overview of the data generation pipeline of MIRIAD}. From top to bottom the MIRIAD processing pipeline is shown. For dataset generation, 894,352 full-text articles from the medical literature were prepared for LLM processing by partitioning the articles into passage chunks of up to 1,000 text tokens. From each passage, an LLM generated medically-related question-answer (QA) pairs, grounded in the passage. In a second step, a sequence of quality control measures was implemented to ascertain relevance, factuality, and groundedness of the semi-synthetically generated QA pairs. This processing pipeline brings forth MIRIAD, a literature-scale and hierarchical resource for operationalized medical knowledge that contains nearly 6 million medical topically organized QA pairs, each grounded in and linked to peer-reviewed literature. MIRIAD enables a range of down-stream use cases, as shown in the bottom row, including serving as a corpus of external knowledge for retrieval-augmented generation (RAG), a supervised dataset for training medical information retrievers, as well as advanced interfaces for users to visually explore, search, and navigate a structured landscape of medical queries and responses, with clickable follow-up literature.}
    \label{fig:overview}
\end{figure}

In this study, we present MIRIAD, a massive corpus of operationalized medical knowledge, i.e., a large-scale collection of medical queries and responses, each pair grounded in the peer-reviewed literature. To construct MIRIAD, we leveraged the Semantic Scholar Open Research Corpus (S2ORC)\cite{lo2020s2orc}. We filtered for articles tagged under “Medicine”, yielding a focused pool of 2,503,836 medical papers. We leveraged the first 894,352 papers for further LLM processing, leaving open the option to further scale up the dataset in future versions. Each article was segmented into passages of up to 1,000 tokens. Each passage was processed by the GPT-3.5-Turbo language model using a structured prompt and detailed instructions to generate self-contained question-answer (QA) pairs. The questions focused on medical and biomedical facts that could be answered solely with the information provided in the source passage, in order to avoid overly general or irrelevant content. Fig.~\ref{fig:overview} provides a step-by-step overview.

The initial generation step resulted in over 10 million raw query-response (QA) pairs, which laid the foundation for the MIRIAD dataset. After subsequent quality control (more details in the next paragraph), the final MIRIAD release consists of 5,821,948 (MIRIAD-5.8M) high-quality question-answering pairs, each linked to a source passage chunk and peer-reviewed paper. Moreover, we also release a refined dataset (MIRIAD-4.4M) where additional quality control steps were implemented. MIRIAD spans 56 medical topics and disciplines, providing a comprehensive view of medical and biomedical knowledge.

\subsubsection*{Multi-step quality control}

To ensure the reliability and domain relevance of the generated QA pairs, we applied a multi-stage filtering pipeline combining automated heuristics, LLM-based supervision, and human expert annotation (Fig.~\ref{fig:overview}, quality control panel).

As a first step, a rule-based filter was designed to eliminate QA pairs that relied on meta-linguistic references to the source passage. Questions that explicitly referred to the passage itself—e.g., those containing phrases such as “the passage states that” or “according to the passage”—were removed using a targeted keyword and phrase list. This step eliminated nearly 5 million low-utility QA pairs, resulting in a filtered set of 5,821,948 QAs across 2,303,282 passages from 812,384 papers.

Next, we applied LLM-based annotation to a small sample of 15,000 QA pairs, instructing GPT-4 to assess each example along two axes: factual correctness and domain relevance. These labels were used to fine-tune a Mistral-7B classifier, which was then applied to the full filtered set. This classifier achieved 81.8\% recall in detecting GPT-4-flagged low-quality examples and produced the MIRIAD-4.4M dataset with 4,487,542 high-confidence QA pairs.

\begin{figure}
    \centering
    \includegraphics[width=1\linewidth]{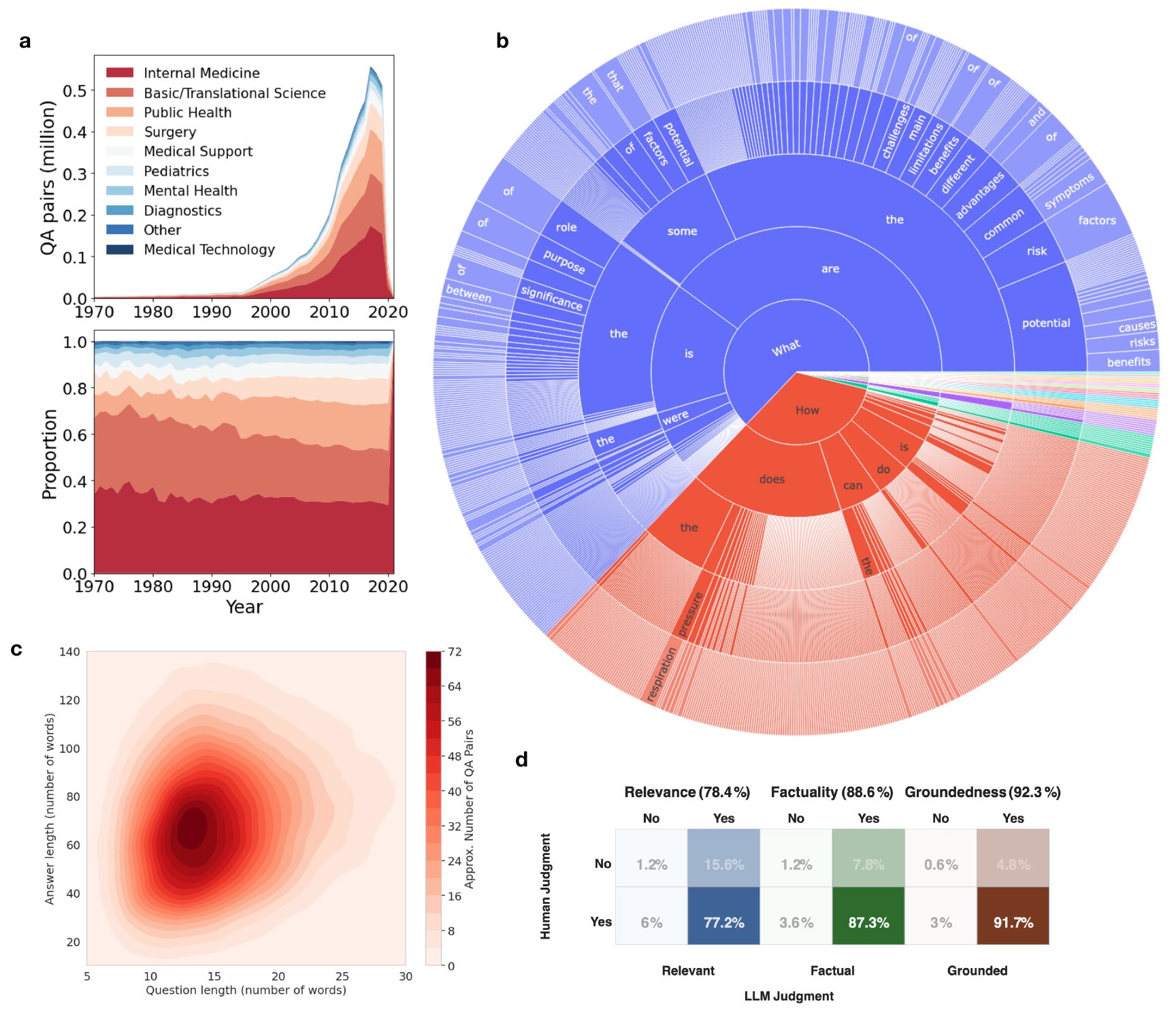}
    \caption{\textbf{MIRIAD data characteristics}. \textbf{a}, Temporal distribution of discipline-specific question–answer (QA) pairs in the MIRIAD dataset. The dataset spans 56 distinct medical domains, which have been grouped into 10 broader categories for clarity in this visualization. This figure illustrates the relationship between the medical domain of each QA pair and the publication year of the corresponding source paper. The top panel shows absolute counts of QA pairs within a given year, the bottom row shows the relative proportion of fine-grained topics within a given year bucket. The 56 fine-grained discipline categories of MIRIAD were collapsed into 10 coarse categories in a two-step process.  First, an LLM processed the categories to produce a candidate mapping. A clinical expert then edited and revised the mapping. For a fine-grained version of the same plot, refer to Fig.~\ref{fig:sup_fig_fine_grained_disciplines} showing all 56 categories.  \textbf{b}, The first five words of a small subset of 1000 MIRIAD questions are visualized hierarchically in a sunburst plot. Color indicates that the different questions share the same first word, e.g., “What” or “How”. \textbf{c}, A kernel density estimate (KDE) plot over the question-and-answer length in MIRIAD is shown. The KDE plot shows that most questions contain less than 30 words and answers less than 140 words, respectively. \textbf{d}, Human annotation results are shown in confusion matrices for the metrics relevance, factuality, and groundedness. Percentages indicate where the LLM judge agreed with the human experts. For example, in 91.7\% of the reviewed samples both human and LLM rate the QA pair as grounded in the source passage, whereas in 0.6\% of the samples both agreed that the sample was not grounded, totaling in 92.3\% agreement.}
    \label{fig:data-characteristics}
\end{figure}

To evaluate the agreement between LLM-based and human annotation, we conducted a human evaluation study with five medical experts to review a subset of 56 passages and 168 QA pairs, with overlapping examples to assess consistency. QA pairs were rated for relevance, factuality, and groundedness. We observed strong agreement between GPT-4 and expert judgments: 92.3\% on groundedness, 88.6\% on factuality, and 78.4\% on relevance (Fig.~\subfigref{fig:data-characteristics}{d}). 

These findings validate the use of LLM-supervised filtering as a scalable alternative to complement expert annotation.

\subsubsection*{Dataset characteristics}

MIRIAD represents one of the largest and most diverse collections of medical query-response pairs to date, comprising 5,821,948 high-quality QA pairs derived from 812,384 peer-reviewed medical papers~(after filtering) across 56 distinct disciplines and topics. Fig.~\subfigref{fig:data-characteristics}{a} illustrates the temporal distribution of QA pairs across disciplines, with both absolute count (top) and relative proportion (bottom) demonstrating comprehensive coverage. A more fine-grained version of the schematic is provided in Fig.~\ref{fig:sup_fig_fine_grained_disciplines}. Fig.~\subfigref{fig:data-characteristics}{a} shows that consistent across time roughly 80\% of the MIRIAD data pertains to internal medicine, basic science, public health, and surgery. The vast majority (98.4\%) of the MIRIAD source text stems from literature spanning the years 1970 to 2021. Fig.~\subfigref{fig:data-characteristics}{b} shows the hierarchical distribution of MIRIAD question word compositions in a sunburst visualization. For improved readability, this graph was computed from a sample of 1,000 QA pairs, where it already reveals a broad range of question formulations.

MIRIAD exhibits strong quality metrics. Human-in-the-loop data quality control via medical expert assessment (Fig.~\subfigref{fig:data-characteristics}{d}) indicates overall high quality of MIRIAD's QA pairs, with strong agreement between human experts and LLM judgments: 92.3\% on groundedness, 88.6\% on factuality, and 78.4\% on relevance. Furthermore, a LLM-based classifier used for quality filtering achieved 81.8\% recall in detecting low-quality examples. When compared to existing medical QA datasets like PubMedQA (211,269 questions), MedMCQA (193,155 questions), and MedQA (61,097 questions), MIRIAD offers an order of magnitude more instruction-response pairs~\cite{jin2019pubmedqa, pal2022medmcqa, jin2021disease}. More importantly, MIRIAD distinguishes itself from traditional medical corpora such as PubMed\footnote{\url{https://pubmed.ncbi.nlm.nih.gov/}}, StatPearls\footnote{\url{https://www.statpearls.com/}}, and textbooks by providing information in an operationalized QA format rather than as unstructured text, which we hypothesized will facilitate more precise retrieval for downstream applications.

Fig.~\subfigref{fig:data-characteristics}{c} shows a heatmap of the question-and-answer lengths in number of words; questions typically range between 15 and 20 words, whereas answers span 60 to 80 words. This shows that MIRIAD contains concise responses that capture essential medical knowledge without unnecessary verbosity. If further verbosity is required, each QA pair's explicit grounding in source text enables seamless back traceability to peer-reviewed literature—a critical feature for clinical applications where evidence provenance is essential (more details in Section~\ref{sec:atlas}).

These characteristics collectively position MIRIAD as a comprehensive knowledge base for downstream AI-based use cases that presuppose accurate, diverse, and well-structured medical information.

\subsection{Interactive MIRIAD atlas} \label{sec:atlas}
To make the MIRIAD broadly available as an interactive resource, we developed MIRIAD-Atlas, a web-hosted user interface as depicted in Fig.~\ref{fig:datamap-atlas}. For this, the MIRIAD dataset is visualized as an interactive point cloud that users can navigate and explore for in-depth medical knowledge. Each point in this UMAP visualization represents a QA pair, with semantically similar content grouped together. First, we embedded each QA pair using sentence-transformers/all-MiniLM-L6-v2 to generate 384-dimensional vectors, then applied sequential dimension reduction via PCA (384→50 dimensions) followed by UMAP (50→2 dimensions) to create the 2D visualization.

The resulting map reveals natural clustering by medical disciplines, with related specialties positioned in proximity to one another. Major clinical specialties such as cardiology, oncology, neurology, and gastroenterology form distinct regions, while cross-disciplinary topics create bridges between specialty clusters. The visualization supports dynamic interaction, allowing users to search for terms through keyword search such as "heart," "cancer," or more specific entities like "TP53" and "HER2," with matching QA pairs highlighted within their semantic context. Users can explore rare conditions like Creutzfeldt-Jakob disease by locating relevant information within the broader medical knowledge landscape.

Source provenance is accessible by hovering over points to view QA content, publication venue, and paper ID, whereas clicking on the data point directly redirects the user to the peer-reviewed source documents for verification and further review.

This interactive interface transforms MIRIAD from a static asset into an exploratory tool for researchers or medical caregivers. It enables rapid knowledge retrieval while maintaining the critical connection to more verbose peer-reviewed sources to back up concise factoids with more in-depth evidence.

\subsection{Retrieval-augmented generation (RAG) with MIRIAD} \label{sec:rag-results}
We systematically evaluated the effectiveness of the MIRIAD corpus as an external knowledge base for retrieval-augmented generation (RAG) tasks, specifically in medical question-answering settings. Additional background and prompts used regarding RAG in medical applications is provided in Supplementary Information \ref{Supp:rag-details}.
For this experiment, we implemented a minimalist RAG architecture to isolate the specific contribution of the MIRIAD corpus. This approach avoids confounding factors associated with advanced retrieval strategies such as hybrid retrieval,  iterative query rewriting or retrieval reranking. Two distinct embedding models were employed to encode user queries and the MIRIAD corpus: sentence-transformers/all-MiniLM-L6-v2 (all-MiniLM) \cite{wang2020minilm}, optimized for general semantic similarity, and BAAI/bge-large-en-v1.5 (BGE-Large) \cite{xiao2024c}, specifically trained for retrieval tasks.

\begin{figure}
    \centering
    \includegraphics[width=1\linewidth]{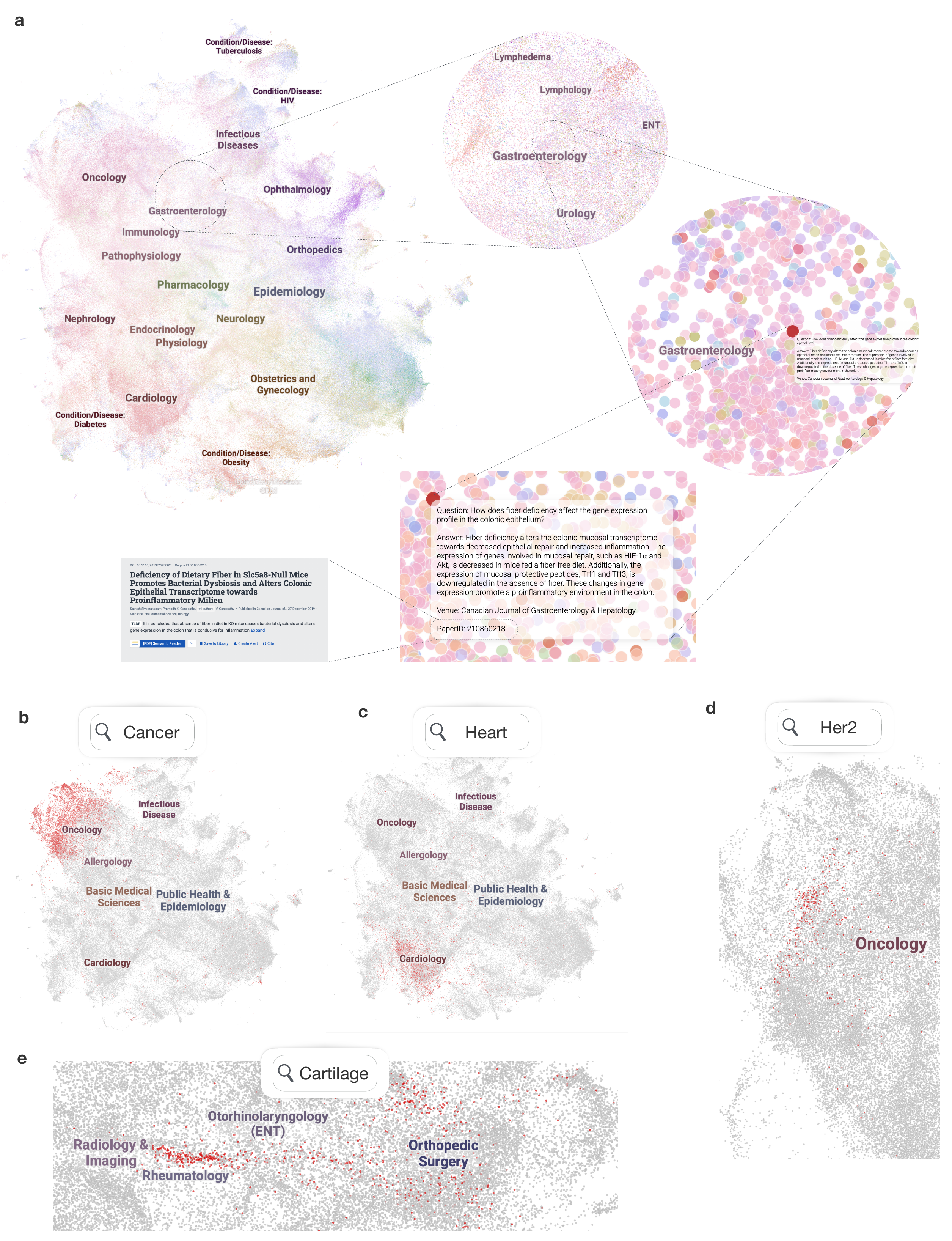}
\end{figure}

\newpage

\begin{center}
  \captionof{figure}{\textbf{Overview of the MIRIAD atlas}. \textbf{a}, MIRIAD-Atlas is an interactive user interface for visually exploring MIRIAD. Each data point represents a question-answer (QA) pair that is grounded in an article from the literature, and is linked to the source passage and metadata (such as the venue, or the date of publication). To facilitate visualization, a subset of 1M data points are displayed. Each data point was classified into one of 56 categories that represent topics and medical disciplines. The data points are visualized in 2D using UMAP based on the text embedding of the QA pairs, respectively. On the right side, enlargements of a small region of the UMAP embedding landscape of MIRIAD is shown. Panels \textbf{b-e} depict how MIRIAD-Atlas enables keyword searches to highlight relevant subsets of MIRIAD, as demonstrated for broader keywords like “heart” or “cancer”, but also more specialized ones: “Her2” appears as a subspace of Oncology, whereas the keyword “cartilage” is spread across different disciplines that relate to it. The smaller visualizations of panels b-e leverage a sample of 300K points.}
  \label{fig:datamap-atlas}
\end{center}
\vspace{-2em}

Text embeddings were indexed within a Qdrant vector database \cite{qdrant2022} using cosine similarity to facilitate retrieval of the relevant entries. To ensure a fair comparison between different RAG corpora, we adopted a fixed context budget approach: retrieved entries were accumulated until the total context length reached a predefined token limit (primarily n=1000; extended results for n=200–2600 are provided in Supplementary Section~\ref{supp:rag-tokens}). Retrieved items—either structured QA pairs from MIRIAD (RAG-MIRIAD) or unstructured literature passages (RAG-Passage)—were appended to the input query to form the augmented prompt, which was then passed to the language model for answer generation. 

We evaluated model performance on the MedMCQA benchmark, a widely recognized and challenging dataset comprising multiple-choice medical questions. Three experimental conditions were compared: retrieval using MIRIAD’s QA pairs (RAG-MIRIAD), retrieval from raw passages (RAG-Passage), and a baseline without retrieval augmentation (No-RAG), where the LLM directly answers the question. 

\begin{table}[h]
    \centering
    \resizebox{\textwidth}{!}{%
        \begin{tabular}{ccccc}
        \toprule
         \multirow{1}{*}{\textbf{Embed Model}} & \multirow{1}{*}{\textbf{Backbone LLM}} &  \textbf{\textit{RAG-MIRIAD}} & \textbf{\textit{RAG-Passage}} & \textbf{\textit{No-RAG}} \\
        \midrule 
             \multirow{3}{*}{bge-large-en-v1.5} & Mixtral-8x7B-Instruct & \textbf{59.69 [58.26, 61.2]} & 55.92 [54.46, 57.45] & 53.48 [52.0, 55.01]\\
            & Llama-3.1-8B-Instruct & \textbf{59.36 [57.88, 60.87]} & 55.94 [54.48, 57.47] & 56.08 [54.63, 57.61]\\
            & claude-3-5-sonnet & \textbf{77.22 [75.9, 78.46]} & 74.64 [73.32, 75.95] & 76.81 [75.47, 78.08]\\
        \midrule
         \multirow{3}{*}{all-MiniLM-L6-v2} & Mixtral-8x7B-Instruct & \textbf{56.37 [54.89, 57.85]} & 54.91 [53.45, 56.40] & 53.48 [52.0, 55.01]\\
            & Llama-3.1-8B-Instruct & \textbf{56.11 [54.67, 57.64]} & 53.07 [51.59, 54.58] & 56.08 [54.63, 57.61]\\
            & claude-3-5-sonnet  & 76.40 [75.07, 77.72] & 75.52 [74.16, 76.84] & \textbf{76.81 [75.47, 78.08]}\\
        \bottomrule
        \end{tabular}
    }
    \caption{\textbf{Off-the-shelf RAG results on the MedMCQA dataset.} Results are reported as accuracy followed by the 95\% confidence interval in square brackets. We observe that leveraging MIRIAD as an external knowledge corpus (RAG-MIRIAD) improves performance on average by 5.13\% (relative gain) compared to basic RAG on chunks of unstructured text from the same literature data (RAG-Passage). This finding was validated across 6 configurations of LLMs and text embedding models.}
    \label{tab:rag_main}
\end{table}

As shown in Table~\ref{tab:rag_main}, leveraging MIRAD QA pairs to augment the generation (RAG-MIRIAD) consistently outperforms generation augmented by raw literature passage (RAG-Passage) across different experimental configurations. This holds true regardless of the underlying language model or embedding approach, with RAG-QA showing relative improvements of 1.16\% to 6.74\% over RAG-Passage. The advantage of structured retrieval was particularly pronounced with BGE-Large embeddings. For instance, Mixtral-8x7B-Instruct improved from 55.92\% (RAG-Passage) to 59.62\% (RAG-MIRIAD), and Llama-3.1-8B-Instruct improved from 55.94\% to 59.36\%. Claude-3.5-Sonnet achieved the highest absolute accuracy (77.22\%) using RAG-MIRIAD and BGE-Large embeddings, though its relative gain over RAG-Passage was smaller—likely due to its stronger built-in knowledge base.  These results highlight the utility of structured retrieval with MIRIAD QA pairs, particularly for smaller, open-source models where efficient use of limited context windows is critical. By offering information in a more condensed and semantically aligned format, MIRIAD enables models to integrate external knowledge more effectively, thereby narrowing the performance gap relative to much larger proprietary models. Our findings suggest that encoding medical knowledge as instruction–response pairs improves LLM performance by aligning the retrieval and representation format with the interactive, instruction-driven distribution of downstream tasks. We further validated the advantage of RAG using MIRIAD on additional benchmarks (MMLU-Med and MedQA-USMLE) as described in Section~\ref{supp:validation-benchmarks} of the supplements.

To further unpack the results of our main RAG experiment, we conducted a detailed discipline-specific analysis of benefits and harms from RAG across 35 medical discipline categories that each problem of the MedMCQA dataset was classified into. Categorization details can be found in Supplementary Information~\ref{supp:medmcqa-cat}. 

Fig.~\ref{fig:greenredbar} illustrates the impact of topic-wise RAG, comparing both RAG-MIRIAD (panel a) and RAG-Passage (panel b) against the baseline without retrieval augmentation (No-RAG). In each panel, green dashed bars indicate cases that were corrected by RAG, red dashed bars represent those made worse, and solid bars show the net effect (gain or loss) per topic. Panel a demonstrates that RAG-MIRIAD—using structured QA pairs—yields clear net gains across many disciplines, as evidenced by the more prominent solid green bars in panel a, compared to panel b. While some cancellation between gains and harms remains, its extent is limited compared to panel b. In contrast, RAG-Passage (panel b), which relies on raw unstructured text, shows more frequent cancellation between helpful and harmful retrievals, resulting in shorter solid green bars and a weaker overall benefit across most topics. These results highlight that structured retrieval from MIRIAD is not only more effective overall but also more consistent across diverse clinical areas. This reinforces prior observations that MedMCQA is particularly challenging to improve through naive passage retrieval, and underscores the importance of information structure in maximizing the effectiveness of retrieval-augmented generation.

\begin{figure}
    \centering
    \includegraphics[width=0.9\linewidth]{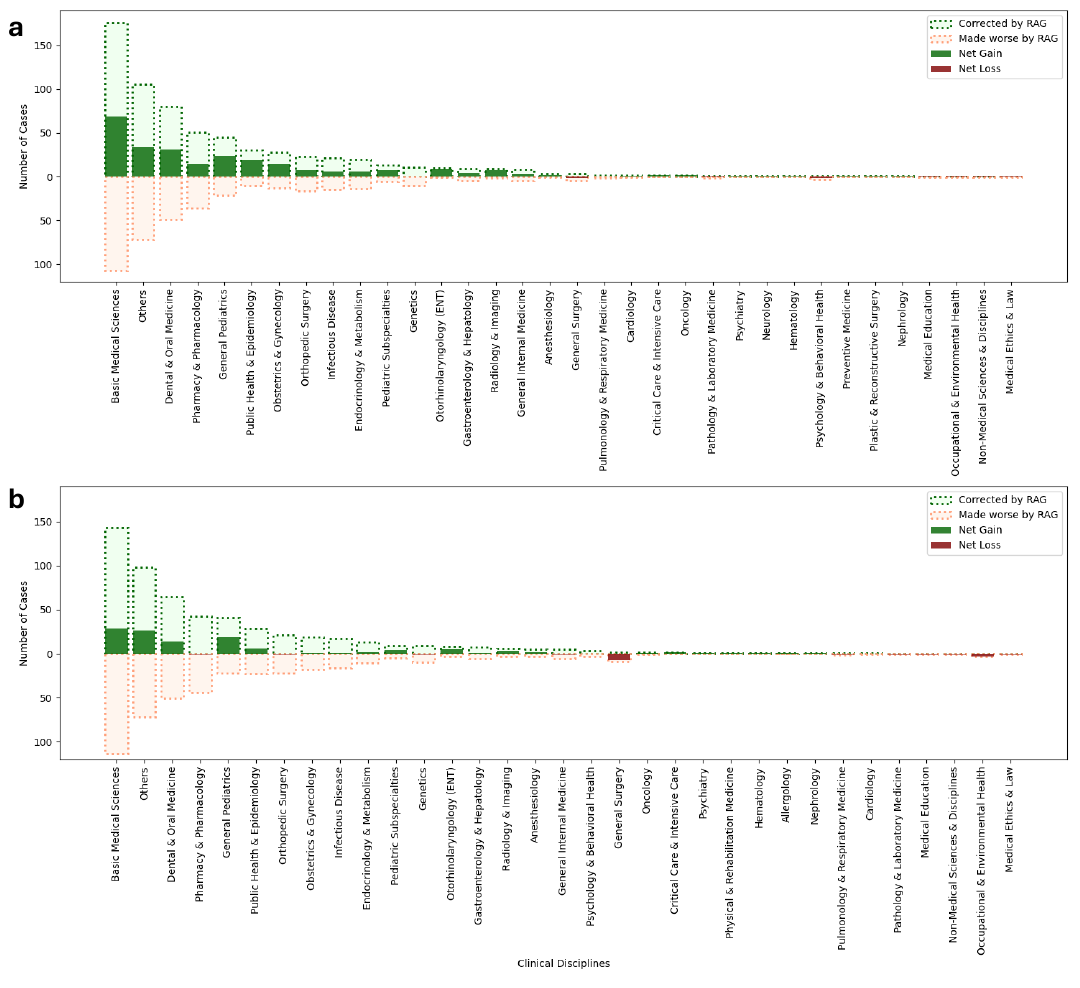}
    \caption{\textbf{Topic-level analysis of RAG performance on MedMCQA}. \textbf{a}, RAG-MIRIAD. \textbf{b}, RAG-Passage. To assess the effect of information structure in retrieval-augmented generation (RAG), we compare RAG-MIRIAD (top panel), which uses concise QA pairs from MIRIAD, with RAG-Passage (bottom panel), which retrieves longer unstructured text chunks from the source literature. In both settings, the total retrieved context is capped at 1000 tokens to ensure a consistent comparison and match the main RAG experiment setup. Each MedMCQA problem was categorized into one of 35 clinical topics using the LLaMA-3-8B-Instruct model. For each topic, bars represent the number of problems that were improved by RAG (green dashed), harmed by RAG (red dashed), and the resulting net effect (solid bars). RAG-MIRIAD exhibits broader and more consistent net benefits, particularly in topics such as basic medical sciences, public health, and dental medicine. In contrast, RAG-Passage shows more cancellation between helpful and harmful retrievals, demonstrating the limitations of conventional passage-based augmentation on this benchmark.}
    \label{fig:greenredbar}
\end{figure}

We further dissected RAG performance at the individual sample level through a controlled attribution analysis to examine the isolated impact of individual retrieved MIRIAD QA pairs from the top-k-retrieved MIRIAD sample set (for details refer to Section~\ref{sec:exp}). We define beneficial samples as retrieved MIRIAD samples that, when integrated into the RAG context, flip an LLM’s answer in a downstream task from incorrect (without RAG) to correct (with RAG), and detrimental samples as samples that flip an answer from correct (without RAG) to incorrect (with RAG).  We use the Jaccard index to measure the overlap between beneficial samples in two RAG pipeline settings, as well as the overlap between detrimental samples in the same settings. As demonstrated in Extended Data Figure \ref{fig:extend_data_fig_jaccard}, both beneficial and detrimental  sample set overlaps remain low (Jaccard Index $<$ 0.14) across different experimental configurations, suggesting that the identity of helpful or harmful retrieved samples is highly dependent on the setup (e.g. backbone LLM or embedder used). Notably, while we aimed to study the impact of individual RAG samples, these findings reveal no systematically detrimental subset across experimental configurations, which provides empirical support for the robustness and quality of the dataset. \\

\subsection{MIRIAD enables the training of medical retriever models}

\begin{figure}[H]
    \centering
    \includegraphics[width=0.6\linewidth]{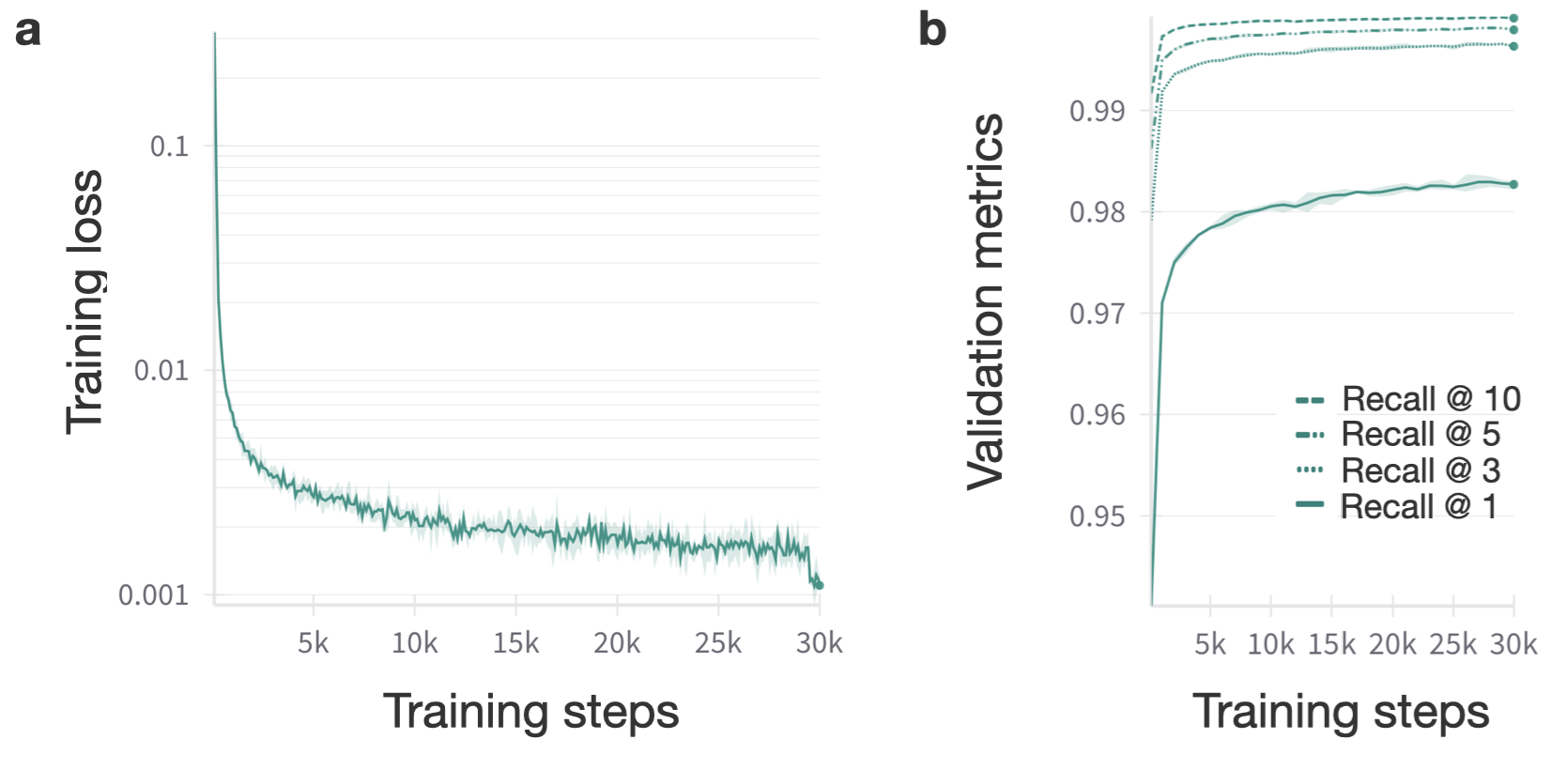}
    \caption{\textbf{Demonstration of retriever training use case}. MIRIAD can be leveraged as a large-scale supervised retrieval dataset to train improved medical information retrieval models. To our knowledge, this has previously not been possible, as large-scale medical retrieval data is lacking or not accessible. Panel a and b show how a retriever can be trained on MIRIAD to produce higher quality embeddings for medical queries and their answers. The curves display the mean; the error bars the standard deviation over 4 repetition runs. \textbf{a}, The training loss of the retriever model is shown on the Y-axis on a log-scale over the course of training (training steps on X-axis). \textbf{b}, This panel shows the retrieval model’s validation metrics for retrieval quality computed on the held-out set of QA pairs. At the baseline of training (x = 0), the metrics for the general-domain retriever is shown (e.g. 0.94 Recall @1).}
    \label{fig:retriever}
\end{figure}

As a proof-of-concept, we investigate whether MIRIAD can be leveraged as a large-scale supervised retrieval dataset to train a retriever model specialized for encoding medical questions and answers in a way that most relevant responses are surfaced by the retriever. To test this ability of the MIRIAD resource, we fine-tuned a widely-used general domain retriever, BAAI/bge-base-1.5, on MIRIAD-5.8M. Fig.~\ref{fig:retriever} shows that MIRIAD is directly amenable to retriever training, where a range of retrieval metrics improve on held-out data over the course of training (here for 30K training steps).

\subsection{MIRIAD helps detecting LLM hallucinations} \label{sec:medhallu-results}

Hallucinations remain a significant barrier to the safe deployment of large language models (LLMs) in clinical practice. To address this, we evaluated the impact of augmenting LLMs with MIRIAD on hallucination detection capabilities in medical contexts. Using the MedHallu benchmark, specifically designed to assess LLMs' ability to distinguish hallucinated answers from accurate ground truth answers, we compared the baseline hallucination detection performance of Llama-3.1-8B-Instruct against the same model augmented with MIRIAD knowledge. Results in Table~\ref{tab:medhallu_rag_comparison5.8M_top3_full} demonstrate that augmentation with MIRIAD significantly improved hallucination detection: the LLM achieved an F1-score of 45.93\% on the full 10k MedHallu dataset, while the LLM with access to MIRIAD reached an F1-score of 68.46\%, reflecting a net improvement of 22.53 points. Notably, evaluation on a higher-quality, human-annotated subset yielded an even greater performance gap, with the MIRIAD-enhanced LLM attaining an F1-score of 65.78\% compared to the No-RAG 28.76\%, representing a substantial absolute improvement of 37.02 percentage points. These results underscore MIRIAD’s effectiveness in improving the ability of LLMs to detect hallucinations in medical conversational tasks.\\

\begin{table}[ht]
    \centering
    \resizebox{\textwidth}{!}{%
    \begin{tabular}{cccccc}
    \toprule
    \multirow{2}{*}{\textbf{Data Subset}} & \multicolumn{2}{c}{\textbf{LLM+MIRIAD}} & \multicolumn{2}{c}{\textbf{LLM Alone}} & \textbf{$\Delta$ MIRIAD}\\
                           & F1 & Accuracy                          & F1 & Accuracy                          & ($\Delta$ F1)\\
    \midrule
    \multirow{2}{*}{10k Full} & \multirow{2}{*}{68.46 [67.78, 69.12]} & \multirow{2}{*}{66.58 [65.93, 67.17]} & \multirow{2}{*}{45.93 [44.82, 46.94]} & \multirow{2}{*}{63.49 [62.83, 64.17]} & \multirow{2}{*}{+22.53}\\
    & & & & & \\
    \multirow{2}{*}{1k Human-Annotated} & \multirow{2}{*}{65.78 [63.42, 67.98]} & \multirow{2}{*}{63.95 [61.80, 66.00]} & \multirow{2}{*}{28.76 [25.51, 32.13]} & \multirow{2}{*}{57.40 [55.35, 59.50]} & \multirow{2}{*}{+37.02}\\
    & & & & & \\
    \bottomrule
    \end{tabular}
    }
    \caption{\textbf{RAG results on the MedHallu benchmark.} Results are reported as F1 score and accuracy followed by the 95\% confidence interval in square brackets. Here we use Llama-3.1-8B-instruct as the backbone model and bge-large-en-v1.5 as the embedding model. Results demonstrating the effectiveness of augmenting the Llama-3.1-8B-Instruct LLM with MIRIAD on the MedHallu benchmark. The table compares the hallucination detection performance, measured by F1-score and accuracy, of the baseline LLM (no additional knowledge) against the same LLM enhanced with MIRIAD knowledge (LLM+MIRIAD). Evaluation is performed on the full 10k MedHallu dataset and a higher-quality human-annotated 1k subset.}
    \label{tab:medhallu_rag_comparison5.8M_top3_full}
\end{table}
\section{Discussion}

We present MIRIAD, a high-quality and diverse dataset comprising 5.8 million medical question-answer (QA) pairs, each grounded in peer-reviewed medical literature. To make this resource broadly accessible to end users (such as researchers, clinicians, or engaged patients), we created an interactive atlas allowing users to explore and navigate an operationalized knowledge landscape and perform keyword-based searches for topics of interest (e.g., specific diseases, drugs, or symptoms). A rigorous quality control regime was implemented through a multi-tiered approach, integrating automated filtering, large language model (LLM) labels, smaller LLM-based classifiers trained on these labels, and clinician-based expert annotations.

MIRIAD offers practical utility in myriad down-stream medical AI and LLM applications. First, we evaluated its efficacy as a structured knowledge corpus in retrieval-augmented generation (RAG) settings for advanced medical question answering. We found that structured knowledge from MIRIAD reduces detrimental retrieval cases compared to traditional RAG methods that leverage the same source text, albeit in an unstructured format. Moreover, MIRIAD can effectively serve as an open-access training dataset of unprecedented scale for domain-specific medical retrieval models. We expect that this will significantly advance medical retriever embedding models which could inform various clinical applications such as digital twins, chat assistants, trial recruitment.

MIRIAD addresses key limitations observed in previous medical datasets. It is the first medical instruction-response dataset at the literature-scale, an order of magnitude larger and more comprehensive compared to prior large medical QA datasets \cite{jin2019pubmedqa, pal2022medmcqa, jin2021disease}. By contrast, previous literature-scale medical corpora primarily consisted of unstructured text, which we found can pose challenges when augmenting LLMs in medical RAG tasks. 

Our findings underscore the value of structured medical knowledge in the form of massive-scale, carefully curated, and literature-grounded question-answer databases. MIRIAD enhances the quality and reliability of medical RAG systems. Even when the underlying source content remains constant, the structured representation of MIRIAD reduces harmful retrieval outcomes by 14.3\%. We expect that MIRIAD-Atlas will make medical knowledge more broadly accessible by letting users world-wide intuitively navigate a landscape of topically grouped question-answer pairs. We found that MIRIAD can out-of-the-box be used to detect medical LLM hallucinations. Consequently, MIRIAD sets a benchmark for future medical knowledge base construction and serves as an accessible, operationalized, and highly interactive resource for medical knowledge, designed to seamlessly interface with LLMs as well as human practitioners.

Despite extensive efforts, MIRIAD does not yet encompass the full breadth of medical knowledge. Rather, it represents a foundational stepping stone rather than a comprehensive endpoint. Furthermore, semi-synthetic QA generation, although subjected to thorough quality control measures, inevitably includes some non-reducible inaccuracies, while we put extensive efforts to minimize them at scale. Accounting for these limitations, and for sake of transparency and flexibility, we release two dataset versions, MIRIAD-5.8M and MIRIAD-4.4M.

Future research should aim to broaden MIRIAD's coverage of medical specialties, subdomains, and emerging areas of clinical knowledge. Refining the QA generation and validation processes will be critical in further minimizing residual inaccuracies.
Future development could enable human-in-the-loop RAG capabilities, where retrieved data points can be dynamically highlighted and filtered based on relevance judgments, empowering clinical users to refine information selection before LLM response synthesis, which could improve performance and accuracy in medical question answering systems. Additionally, exploring more advanced retrieval architectures and integration strategies will likely unlock even greater utility from structured datasets like MIRIAD, pushing forward the capabilities of medical chat applications. 

\bibliographystyle{plainnat}
\bibliography{biblatex-miriad}


\section{Methods}
\subsection{Dataset generation}
We used the S2ORC dataset, a large-scale, open-access corpus of scientific articles, each accompanied by metadata, such as document categories \cite{lo2020s2orc}. To isolate medical literature, we filtered the S2ORC metadata for papers that had “Medicine” listed as one of their categories, which reduced the dataset to 2,503,836  medical articles, of which we leveraged the first 894,352 articles for LLM processing. 

Each paper was segmented into passages of at most 1,000 tokens. Rather than splitting the text at the word level, which risks producing fragmented sentences, we split each document into full sentences delimited by punctuation. We tokenized each sentence using the tiktoken tokenizer, and accumulated them into a single passage until the total token count reached the 1,000-token limit. If the next sentence would cause the passage to exceed this limit, the current passage was finalized. If the remaining sentences at the end of a paper could not collectively fill a 1,000-token chunk, they were still included as a final, shorter passage.

Since we split passages at the sentence level, we also applied a filtering step to discard excessively long sentences after observing that they often contained noisy or malformed content. We conducted a manual review by randomly sampling 100 sentences from each of four token-length bins: 200–300, 300–400, 400–500, and 500–600 tokens. We found that the proportion of irrelevant sentences increased with length: 15\% in the 200–300 token range, 35\% in 300–400, 48\% in 400–500, and 67\% in 500–600. Based on these observations, we excluded sentences exceeding 400 tokens from our final dataset.

For the purpose of generating a large-scale retrieval dataset that spans a wide range of medical fields, we employed a highly structured prompt to guide the LLM's generation. This prompt is meant to ensure that the LLM does not produce overly general questions that are not directly answerable from the passage and that the questions are neither too general nor referred to specific figures or studies in the passages.

\subsection{Quality control}
\subsubsection*{Automated filtering}
After processing 894,352 academic papers, we generated 10,677,724 raw question answer pairs with 3,560,470 passages. Then, we ran an automated keyword-based filtering procedure to remove QA pairs in which the question explicitly referenced the passage. Specifically, we removed any question or answer containing phrases of the form “the passage” or “the study”. Additional rule-based filtering was applied, which is described in  Supplementary Information \ref{supp:add-info-quality-control}.

After this stage, we were left with 5,821,948 question answer pairs with 2,303,282 passages from 812,384 academic papers.

\subsubsection*{LLM-annotation}
We randomly selected 15,000 questions from the original unfiltered dataset and used GPT-4-0613 to generate supervised quality control labels. We focused on the following metrics
\begin{enumerate}
    \item Factual Accuracy: The answers should be factually correct and accurate.
    \item Relevance: The Q\&A should refer to medically relevant content.
\end{enumerate}

\subsubsection*{Classifier training based on LLM annotation}
We finetuned Mistral-7B-Instruct-v0.2 on the GPT-4 labels and were able to filter the 5,821,948 question answer pairs further down to 4,487,542. To validate the classifier's results, we ran an experiment to correlate its responses with GPT-4. We observed 81.8\% recall and 69.7\% precision, which means that 81.8\% of the GPT-4 labeled bad examples were correctly identified as low quality and 69.7\% of the examples were correctly identified as low quality by the classifier. In absolute terms, this means that out of the 1,334,480 examples identified as low-quality, we estimate that as many as 400,000 examples (30\% of the discarded examples) were incorrectly discarded in MIRIAD-4.4M.

\subsubsection*{Human annotation}
We developed a web-hosted, credentialed human evaluation app (see Supplementary Information, Fig.~\ref{fig:sup_fig_streamlit_annot_app}) using “Streamlit” and engaged five medical experts to evaluate a combined total of 56 passages and 168 question-answer pairs associated with them.  To assess inter-annotator agreement, 5 out of the 56 passages were evaluated by all the experts (corresponding to 15 question-answer pairs). After iterative user experience improvements, the final app first displayed question-answer pairs followed by the corresponding passage, so that experts had initial context before reading the lengthier passage. The annotations were collected in a multiple-choice format, where experts could indicate whether a given question-answer pair was factual, medically relevant, grounded in the passage, or any combination of the three. 

Across the 15 overlapping question-answer pairs, the agreement proportions where all annotators unanimously agreed were 73.3\% for relevance, 46.7\% for factuality, and 60\% for groundedness. We also evaluated the agreement between human judgments and GPT-4 on the same set of pairs, which was 93.3\% for relevance, 76.0\% for factuality, and 81.3\% for groundedness.

\subsection{MIRIAD atlas creation}
To create MIRIAD-Atlas, the MIRIAD dataset was visualized as an interactive point cloud that users can navigate and explore in the web browser for in-depth medical knowledge. First, we encoded each QA pair of MIRIAD using sentence-transformers/all-MiniLM-L6-v2 into 384-dimensional real-valued vectors, then applied sequential dimension reduction via PCA (384→50 dimensions) followed by UMAP (50→2 dimensions) to create the 2D visualization. As a result, each point in MIRIAD-Atlas represents a QA pair, with semantically similar content grouped together.

To categorize each sample (i.e., QA pair) of MIRIAD, we processed each sample with  Llama-3-8b-Instruct using a structured prompt to assign one of 50+ topics discipline-related categories. Then we further map the llama3 identified categories into 56 unified and stabilized disciplines (details in S5) with guidance from medical experts. The interactive visualization was built with the help of the visualization engine provided by the Python package “datamapplot 0.5.1”. 

\subsection{Experiments} \label{sec:exp}

\subsubsection*{Basic RAG experiment with MIRIAD}
We evaluated the effectiveness of MIRIAD to serve as an external knowledge corpus in retrieval-augmented generation (RAG) tasks, specifically in advanced medical question-answering problems. Additional background regarding RAG in medical applications is provided in Supplementary Information \ref{Supp:rag-details}.

We implemented a minimalist RAG architecture to isolate the specific contribution of the MIRIAD corpus. This approach avoids confounding factors associated with advanced retrieval strategies such as hybrid retrieval \cite{omrani2024hybrid, li2024retrieval},  iterative query rewriting \cite{liu2024query} or retrieval reranking \cite{mao2024rafe}. Two distinct text embedding models were used to encode user queries and the QA pairs of the MIRIAD corpus: “sentence-transformers/all-MiniLM-L6-v2” \cite{wang2020minilm} (denoted as all-MiniLM), optimized for general semantic similarity, and “BAAI/bge-large-en-v1.5” \cite{xiao2024c} ( denoted as BGE-Large), which was specifically trained for retrieval tasks. Text embeddings were indexed within a Qdrant vector database \cite{qdrant2022} using cosine similarity to facilitate retrieval of the relevant entries. To ensure a fair comparison between different RAG corpus, we adopted a fixed context budget approach: retrieved entries were accumulated until the total context length reached a predefined token limit (primarily n=1000; extended results for n=200 to 2600 are provided in Supplementary Section~\ref{supp:rag-tokens}). Retrieved items, either structured QA pairs or raw passage texts, were integrated to the user query contexts, and subsequently passed to the generative language models. As for language model backbones, we employed open source models Mixtral-8x7B-instruct, and Llama-3.1-8B-instruct which were run locally  on 8 H200 GPU Servers, as well as claude-3-5-sonnet-20241022 which was inferenced using API calls for answer generation.

For RAG performance evaluation, we used the MedMCQA benchmark \cite{pal2022medmcqa}, a widely recognized and challenging benchmark comprising multiple-choice medical questions \cite{xiong2024benchmarking}. Three experimental conditions were compared: retrieval using MIRIAD’s QA pairs (RAG-MIRIAD), retrieval from raw passages (RAG-Passage), and a baseline without retrieval augmentation (No-RAG), where the LLM directly answers the question. The employed prompts for this experiment can be found in Supplementary Information \ref{supp:rag-prompts}. 

To further probe the robustness of the structured retrieval advantage and assess whether this advantage generalizes beyond MedMCQA, we extended our evaluation to two additional medical QA benchmarks—MMLU-Med and MedQA-USMLE—under the same fixed context budget approach (n=1000). We focused on using BGE-Large as the embedder, which is with larger capacity and stronger retrieval alignment than all-MiniLM. Results for these datasets, included in the Supplementary Section~\ref{supp:validation-benchmarks}, further support the hypothesis that structured retrieval with MIRIAD enhances performance across diverse medical reasoning tasks. 

\subsubsection*{Discipline-specific impact analysis of retrieval-augmented generation on MedMCQA}
To assess how retrieval-augmented generation (RAG) affects performance across different medical subdomains, we conducted a stratified analysis over the MedMCQA benchmark dataset. Each question in the benchmark dataset was automatically assigned to one of 35 medical discipline categories using the Llama-3-8B-Instruct model. The categorization prompt and procedure are detailed in Supplementary Information \ref{supp:medmcqa-cat}. 

We dissected two different configurations: RAG-Passage and RAG-MIRIAD. For each question, we determined whether the answer under each of these two RAG configurations was corrected by RAG content or made worse by RAG content compared to the No-RAG baseline.

For each discipline, we calculated the number of questions that were positively affected (i.e., correct under RAG but incorrect under No-RAG), negatively affected (i.e., incorrect under RAG but correct under No-RAG), or unaffected. These counts were visualized using bar plots to show helpful and harmful retrieval effects, as well as the net effect.

\subsubsection*{Retrieval sample attribution analysis}
To further examine the effect of individual retrieved samples in RAG with MIRIAD, we conducted a controlled attribution analysis centered on the top-k retrieved MIRIAD samples across varying RAG pipeline configurations (e.g., two different backbone LLMs as well as two embedding models). We adopted the same minimalist RAG pipeline settings as in our prior experiments to ensure consistency. For each downstream question, we isolated each individual retrieved MIRIAD sample from the top-k set (k = 3, 10, 20) and integrated it alone into the RAG context, systematically testing its individual influence. We defined beneficial samples as retrieved samples that, when added in isolation, flipped the LLM’s answer from incorrect (without RAG) to correct (with RAG), and detrimental as samples that flipped an answer from correct (without RAG) to incorrect (with RAG). To quantify the consistency of these retrieval contributions across different configurations, we analyzed the overlap between beneficial sample sets and between detrimental sample sets, visualized via Venn diagrams (Supplementary Information \ref{supp:add-details-rag}). For concise reporting, we focused on overall set similarity, measured using the Jaccard index, defined as the size of the intersection divided by the size of the union between two sets $J(A, B) = \frac{|A\cap B|}{|A\cup B|}$. This metric provides an interpretable measure, with values closer to 1 indicating higher overlap and values near 0 indicating minimal overlap.

\subsubsection*{Retriever training}

For the retriever training experiment, we leveraged MIRIAD-5.8M and splitted the dataset into training, validation, and testing splits (97\%, 1.5\%, 1.5\% each). We leveraged the embedding model “bge-base-en-v1.5” and continually pre-trained it on the train set for 30k steps which roughly corresponds to one full epoch over the training data, or 14.8 hours in wall clock time. For training, a single Nvidia H200 GPU was used at a batch size of 192. The information retrieval (IR) training objective and IR evaluation were computed using the sentence transformers framework. We did not employ any hyperparameter tuning, but ran with standard configurations (learning rate 1e-5, Adam optimizer with standard $\beta$ values of $\beta_1$ = 0.9, and $\beta_2$: 0.999). We ran 4 repetition runs with different random seeds to report the error bands as shown in Fig.~\ref{fig:retriever}. Validation metrics (Fig.~\subfigref{fig:retriever}{b}) are reported on the held-out validation split.

\subsubsection*{Hallucination detection experiment}
To evaluate whether structured domain knowledge from MIRIAD can improve hallucination detection in medical contexts, we leveraged the MedHallu benchmark \cite{pandit2025medhallu}, a dataset specifically constructed to assess LLMs' ability to distinguish between accurate and hallucinated responses. The dataset consists of questions paired with either a ground-truth answer or a hallucinated response.

We adopted the original experimental setup proposed in the MedHallu paper, using Llama-3.1-8B-Instruct as the base model for classification. The task was framed as a binary classification problem: given a question-answer pair, the model outputs whether the answer is factual or hallucinated. We implemented a vanilla retrieval-augmented generation (RAG) pipeline using MIRIAD as the external knowledge corpus. For each input sample, the top-k = 3 most relevant QA pairs from MIRIAD were retrieved using dense retrieval, and concatenated with the input as additional context.

Two configurations were evaluated: (1) the baseline model without any retrieval (LLM Alone), and (2) the same model augmented with retrieved MIRIAD context (LLM + MIRIAD). The model was prompted to make a hallucination detection judgement based on both the input and, if applicable, the retrieved evidence. Performance was measured using F1-score on both the full MedHallu dataset and a higher-quality, human-annotated subset. Prompts used for hallucination detection are provided in Supplementary Information \ref{supp:hallucination_detection}.

\renewcommand{\figurename}{Extended Data Figure}
\setcounter{figure}{0} 

\begin{figure}[h!]
    \centering
    \includegraphics[width=0.9\linewidth]{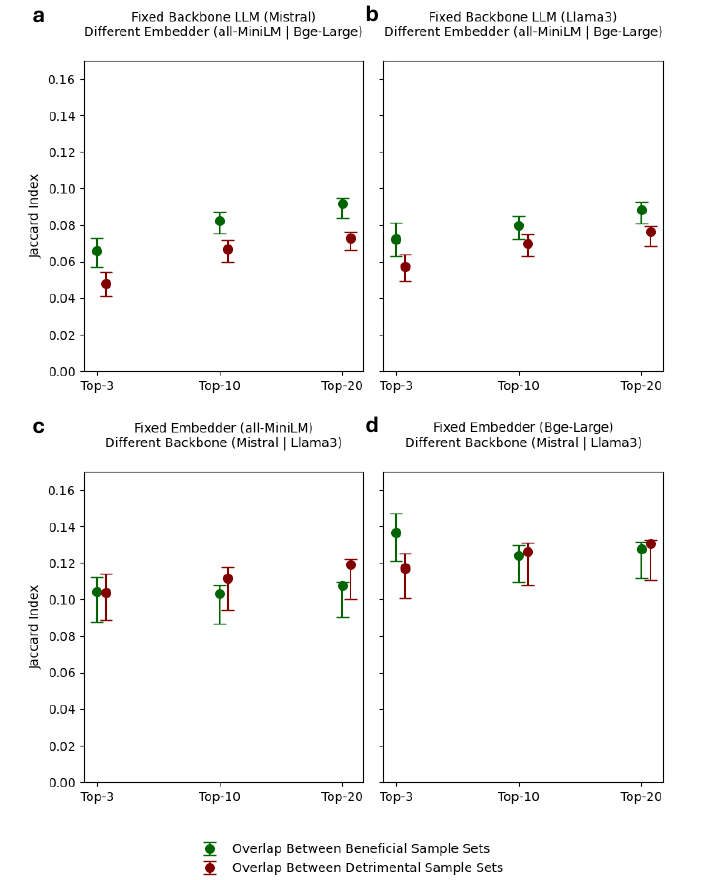}
\end{figure}

\newpage

\begin{center}
  \captionof{figure}{\textbf{Beneficial and detrimental RAG samples do not generalize across different RAG system configurations}. Visualization of Jaccard index that is used to measure the overlap between beneficial sample sets and between detrimental sample sets of different RAG pipeline settings, measured at Top-3, Top-10, and Top-20 retrieved contexts. (\textbf{\textcolor{mygreen}{Beneficial samples}} - samples that improve QA when integrated; \textbf{\textcolor{myred}{Detrimental samples}} - samples that harm QA when integrated). Error bars indicate 95\% confidence intervals. Overall, the overlaps between both beneficial sample sets and the overlaps between detrimental sample sets remain low (Jaccard Index $<$ 0.14), suggesting that the identity of helpful or harmful samples is highly specific to the system setup, posing challenges for designing generalizable retrieval strategies. \textbf{a}, Fixed backbone LLM (“Mistral” refers to mixtral-8x7b-instruct) with varying embedder (all-MiniLM vs. BGE-Large). \textbf{b}, Fixed backbone LLM (“Llama3” refers to llama-3.1-8b-instruct) with varying embedder. \textbf{c}, Fixed embedder (all-MiniLM) with varying backbone LLM (Mistral vs. Llama3). d, Fixed embedder (BGE-Large) with varying backbone. In all cases, the low Jaccard index indicates limited overlap between the beneficial or detrimental samples identified under different configurations, underscoring that retrieval effects are highly setting-dependent.}
  \label{fig:extend_data_fig_jaccard}
\end{center}

\subsection*{Data availability}
The MIRIAD dataset, both 5.8M version and 4.4M version, will be made accessible on huggingface (under \url{https://huggingface.co/miriad}) under a ODC-By v1.0 license upon publication. Additional licensing details are provided in Supplementary Section~\ref{sec:licensing}.

\subsection*{Code availability}
The code used in MIRIAD generation, quality control, and downstream experiments has been made publicly available for non-commercial academic use and can be accessed on github \url{https://github.com/eth-medical-ai-lab/MIRIAD}. Pretrained models used for embedding and generation (e.g., LLaMA 3, Mixtral, all-MiniLM, BGE-Large) were accessed via their official Hugging Face repositories, as cited in the Methods and References sections. 

\subsection*{Acknowledgements}
E.J.T. is supported by the NIH/ National Center for Advancing Translational Sciences grant UL1TR001114. We gratefully acknowledge the support of NSF under Nos. OAC-1835598 (CINES), CCF-1918940 (Expeditions), DMS-2327709 (IHBEM), IIS-2403318 (III); Stanford Data Applications Initiative, Wu Tsai Neurosciences Institute, Stanford Institute for Human-Centered AI, Chan Zuckerberg Initiative, Amazon, Genentech, Hitachi, and SAP. The content is solely the responsibility of the authors and does not necessarily represent the official views of the funding entities.

\subsection*{Author contribution statement}
M.M. conceived the study. S.A. and M.M. created the initial version of the datasets. S.A. and M.M. designed the quality control pipeline. Q.Z. and S.R. implemented the retrieval augmented generation (RAG) pipeline. Q.Z. and M.M. conducted the main experiments of this study. Q.Z. created MIRIAD-Atlas. C.Z., S.O., M.P., M.M., and E.R. contributed to the quality control analyses. E.J.T. and J.L. provided valuable feedback throughout the study design and implementation phases. M.M. designed Figure \ref{fig:overview}. M.M., Q.Z., and S.A. designed Figure \ref{fig:data-characteristics}. Q.Z. and M.M. designed Figure \ref{fig:datamap-atlas}. Q.Z. designed Figure \ref{fig:greenredbar}. M.M. designed Figure \ref{fig:retriever}. Q.Z. created Table~\ref{tab:rag_main} and \ref{tab:medhallu_rag_comparison5.8M_top3_full}. M.M., Q.Z., and S.A. created the first draft of the manuscript. All authors contributed to the final version of the manuscript.\\

\appendix 
\newpage

\section*{\LARGE{Supplementary materials}}

\renewcommand{\thesection}{S\arabic{section}} 
\setcounter{section}{0}                       

\makeatletter
\renewcommand{\theHsection}{section.S\arabic{section}}
\makeatother

\section{Additional details for dataset generation} \label{supp:add-details-data-gen}
We employed a structured prompt to guide the LLM’s generation of question–answer pairs, emphasizing two key objectives: groundedness and medical relevance. To ensure groundedness, we required that all questions be directly answerable from the content of the passage. And to ensure medical relevance, we included a comprehensive list of 64 medical categories within the prompt to encourage questions within established domains of medical knowledge while avoiding references to specific studies, tables, or figures. 

\subsection{Dataset generation prompt}

\begin{tcolorbox}[colback=gray!5!white, colframe=gray!89!black, title=MIRIAD Data Generation]
Please create three questions that are directly answerable from the passage's content. It's imperative that these questions do not focus on or refer to any specific studies, figures, or tables mentioned in the passage. Instead, they should encourage a deeper exploration and understanding of the passage's general content and ideas. They should be framed in a way that their answers can be clearly drawn from the context of the passage. They should be from the following categories: \\

Condition/Disease/Treatment/Symptom/Cause/Risk Factors/Prevention/Diagnosis/Prognosis/Pharmacology/Anatomy/Physiology/Biochemistry/Pathophysiology/Epidemiology/Surgical Procedures/Nutrition and Diet/Genetics and Genomics/Pediatrics/Geriatrics/Psychology/Psychiatry/Obstetrics and Gynecology/Dentistry/Immunology/Virology/Environmental and Occupational Health/Pharmacy and Drug Dispensation/Rehabilitation/Microbiology/Endocrinology/Neurology/Radiology/Oncology/Cardiology/Gastroenterology/Dermatology/Nephrology/Ophthalmology/\\Orthopedics/Hematology/Rheumatology/Pulmonology/Urology/Otorhinolaryngology (ENT)/Veterinary Medicine/Addiction Medicine/Chiropractic Medicine/Palliative Care/Bioinformatics/Transplantation/Toxicology/Parasitology/Stem Cell Biology/Podiatry/Hepatology/Sports Medicine/Family Medicine/Sleep Medicine/Critical Care Medicine/Medical Ethics/Forensic Medicine/Infectious Diseases/Emergency Medicine \\

Please structure your responses in the following format: \\

Question 1: {question that prompts a detailed exploration of a central theme or key concept found in the passage. This question should be answerable based solely on the passage's content, without needing to reference specific studies or data.} \\

Answer 1: {an associated in-depth answer grounded in the passage’s context, providing thorough information and explanation}\\

\end{tcolorbox}

\begin{tcolorbox}[colback=gray!5!white, colframe=gray!89!black, title=MIRIAD Data Generation (Cont'd)]
Question 2: {question that prompts a detailed exploration of a central theme or key concept found in the passage. This question should be answerable based solely on the passage's content, without needing to reference specific studies or data.}\\

Answer 2: {an associated in-depth answer grounded in the passage’s context, providing thorough information and explanation}\\

Question 3: {question that prompts a detailed exploration of a central theme or key concept found in the passage. This question should be answerable based solely on the passage's content, without needing to reference specific studies or data.}\\

Answer 3: {an associated in-depth answer grounded in the passage’s context, providing thorough information and explanation}\\

Passage to consider: {insert passage here} \\

Examples of Inappropriate Questions: \\

"How does the study demonstrate the general effectiveness of the approach discussed in the passage?"\\

Why It's Inappropriate: This question inappropriately focuses on a specific "study," contrary to the aim of engaging with the passage's general content.\\

"What are the key findings of the study mentioned in the passage regarding the overall topic?"\\

Why It's Inappropriate: This question incorrectly seeks information from a specific "study," while the goal is to explore the passage's broader themes.\\

"In what ways does the study alter our understanding of the main subject discussed in the passage?"\\

Why It's Inappropriate: This question wrongly inquires about the impact of a specific "study," rather than encouraging engagement with the overall content of the passage.\\

\end{tcolorbox}

\section{Additional details for quality control} \label{supp:add-info-quality-control}
\subsection{Additional rule-based filtering}
Following initial keyword-based filtering, the dataset consisted of 5,822,163 question answer pairs. Among these, 191 had implausible year values (i.e., earlier than 1790 or later than 2024). Rather than excluding these entries, we set the year field to None, as manual inspection confirmed the associated papers were valid. We removed 17 entries in which any of the passage, question, or answer fields were empty. An additional 198 entries were excluded due to malformed, duplicated, or irrelevant content within the question or answer fields. In total, 215 entries were removed, resulting in a final count of 5,821,948 for MIRIAD-5.8M. Of these 215 removed entries, 141 were also present in the MIRIAD-4.4M subset. These were excluded, reducing the final count for MIRIAD-4.4M to 4,487,542 question answer pairs.

\subsection{Relevance prompt}
This prompt asks a language model to classify medical question-answer pairs as “good” or “bad” based on their medical relevance. A good pair conveys clear, generalizable medical knowledge (e.g., treatments or clinical procedures), while a bad pair focuses on study-specific details like experimental design, figures, or results that lack broader biomedical value. 

\begin{tcolorbox}[colback=gray!5!white, colframe=gray!89!black, title=Prompt for Quality Control - Relevance Check]
I took passages of text from the medical literature and converted each passage into Q\&A problems with the goal to distill the medical knowledge of the literature into a set of Q\&A problems. Having generated them, I noticed that some Q\&As are bad because they do not contain medically or biomedically relevant information. \\

Ideally, a Q\&A conveys a relevant piece of medical knowledge like this good example:\\

"Q: What is the first-line therapy for M. tuberculosis pneumonia?\\

A: Rifampin, isoniazid, pyrazinamide, and ethambutol are first-line antitubercular medications."\\

By contrast, a bad Q\&A would have one or more of the following issues:\\

1. Refers to the details of a specific study (that is presented in the passage), such as specific details about the study's experimental design, the used statistical methods used in this study, tables or figures that appear in it, study dates, locations, funding sources, or other details that are not essential for understanding the underlying medical facts.\\

\end{tcolorbox}

\begin{tcolorbox}[colback=gray!5!white, colframe=gray!89!black, title=Prompt for Quality Control - Relevance Check (Cont'd)]

2. Is heavily dependent on study-specific details that cannot be understood without the original passage, such as discussing the study's specific findings, limitations, or conclusions without providing sufficient background information.\\

3. Focuses on experimental methods or protocols that, while medically related, are too specific to the referenced study and do not convey broadly relevant medical knowledge.\\

If the Q\&A does not have any other bad aspects mentioned above, then the Q\&A should be classified as good. A Q\&A that effectively communicates a clinical procedure, treatment approach, or any clinical or biomedical knowledge in a clear and concise manner should be considered good.\\

Use the above criteria to judge whether the following Q\&A is good or bad. Classify the Q\&A as either "good" or "bad" and provide a short explanation for the classification.
\end{tcolorbox}

\subsection{Factuality prompt}
This prompt asks a language model to evaluate whether a question-answer pair is factually correct. A pair is discarded if it contains information that is false, inaccurate, or contradicts established medical knowledge.

\begin{tcolorbox}[colback=gray!5!white, colframe=gray!89!black, title=Prompt for Quality Control - Factuality Check]
I have converted passages of text from medical literature into Q\&A problems with the goal of distilling the medical knowledge of the literature into a set of Q\&A problems. This dataset will be used to train medical language models, and factual accuracy is of utmost importance.\\

However, I noticed that some Q\&As are problematic because they contain factually incorrect information. A Q\&A is considered factually incorrect if it contains information that is false, inaccurate, or contradicts established medical knowledge.\\

Q\&As can be categorized into two types based on their factual accuracy:

1. correct

2. incorrect\\

Task: Using the criteria above, judge whether the Q\&A is correct or incorrect. Provide a short explanation for your classification.
\end{tcolorbox}

\subsection{Groundedness prompt}
This prompt asks the model to determine whether the answer in a question-answer is grounded in the provided passage. A pair is grounded if all answer content can be directly supported by the passage, and it is ungrounded if it includes external information.

\begin{tcolorbox}[colback=gray!5!white, colframe=gray!89!black, title=Prompt for Quality Control - Groundedness Check]
I have converted passages of text from medical literature into Q\&A problems with the goal of distilling the medical knowledge of the literature into a set of Q\&A problems. This dataset will be used to train medical language models, and groundedness is of utmost importance.\\

A Q\&A is considered grounded only if:

1. Every piece of information in the answer can be traced back to the provided passage

2. The answer does not include any information that goes beyond the content of the passage\\

A Q\&A is considered ungrounded if:

1. The answer contains any information not present in the passage

2. The answer makes claims or assertions that cannot be directly supported by the passage

3. The answer extrapolates beyond what is stated in the passage\\

Task: Judge whether the Q\&A is grounded or ungrounded based on the passage provided. Provide a short explanation for your classification.
\end{tcolorbox}
\renewcommand{\figurename}{Figure}
\setcounter{figure}{0}
\renewcommand{\thefigure}{S\arabic{figure}}

\setcounter{table}{0}
\renewcommand{\thetable}{S\arabic{table}}

\makeatletter
\renewcommand{\theHfigure}{figure.S\arabic{figure}}
\renewcommand{\theHtable}{table.S\arabic{table}}
\makeatother

\begin{figure}[H]
    \centering
    \includegraphics[width=0.8\linewidth]{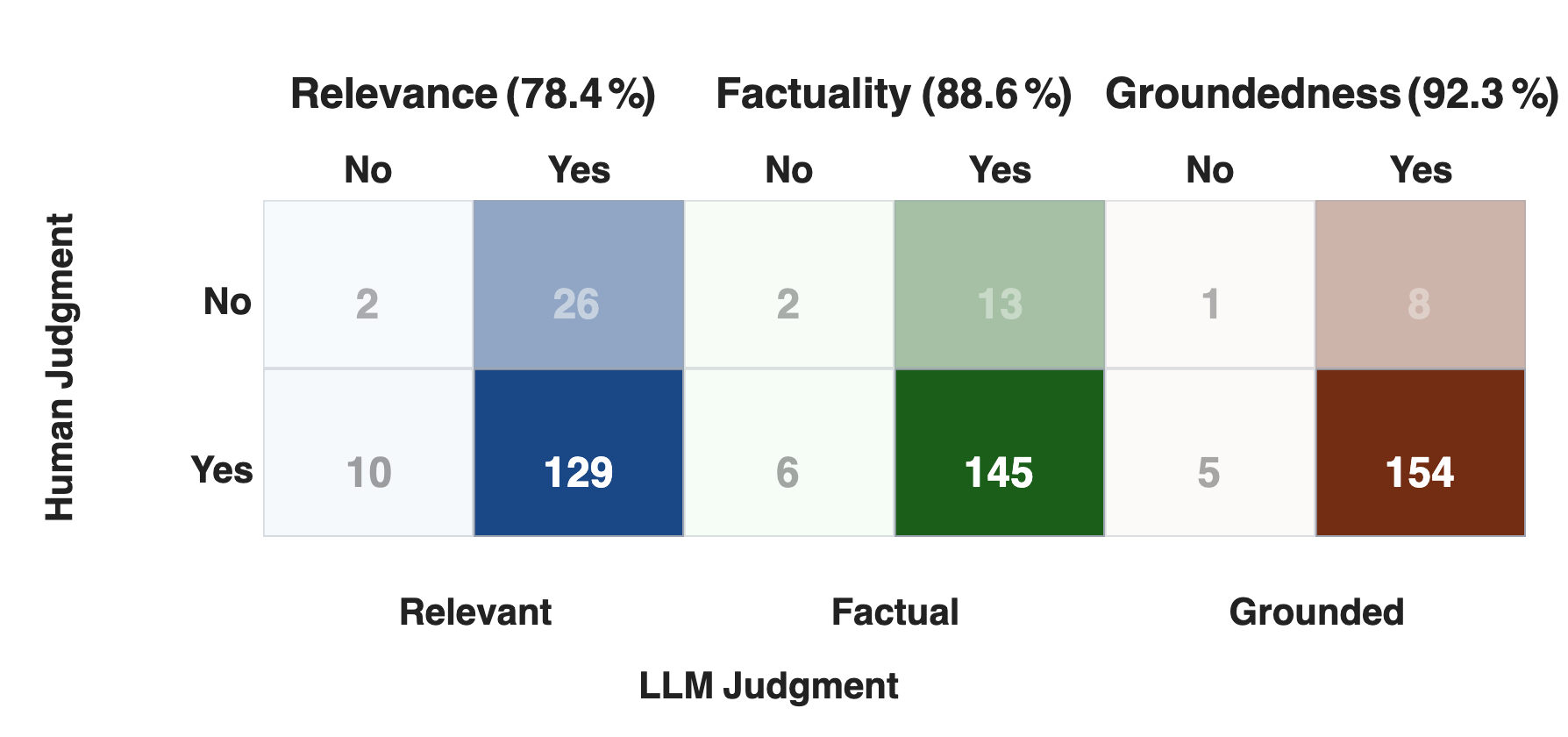}
    \caption{\textbf{Human annotation results}. For three evaluation metrics, we report the absolute counts where the LLM judge agreed or disagreed with human annotation of QA samples.}
    \label{fig:sup_fig_expert_annot}
\end{figure}
\begin{figure}[p]
    \centering
    \includegraphics[width=1\linewidth]{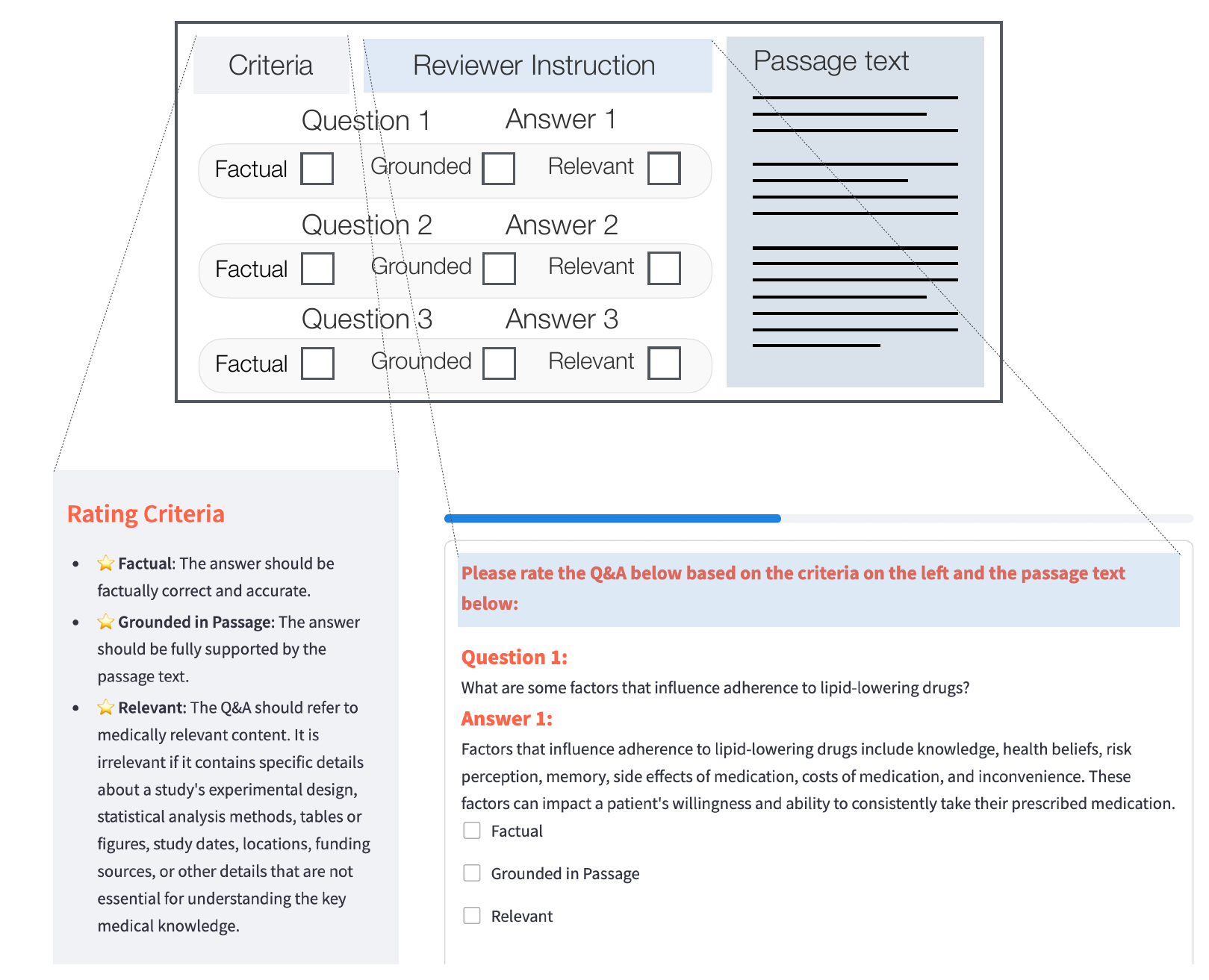}
    \caption{\textbf{Human evaluation app.} Five clinical reviewers were using the evaluation app, above shown in schematic, and below exemplified as a screenshot of the app. Reviewers were instructed, and also had a tab on the left to remind them on the rating criteria even while scrolling down through the passage text. For better user experience, we first present the QA pairs and only then let the reviewer browse through the passage below. As it was challenging for reviewers to assign fine-grained scores, we designed the metrics as tick boxes, where reviewers check whether a given QA pair is factual, grounded in the source passage, and relevant.}
    \label{fig:sup_fig_streamlit_annot_app}
\end{figure}
\begin{figure}[p]
    \centering
    \includegraphics[width=1\linewidth]{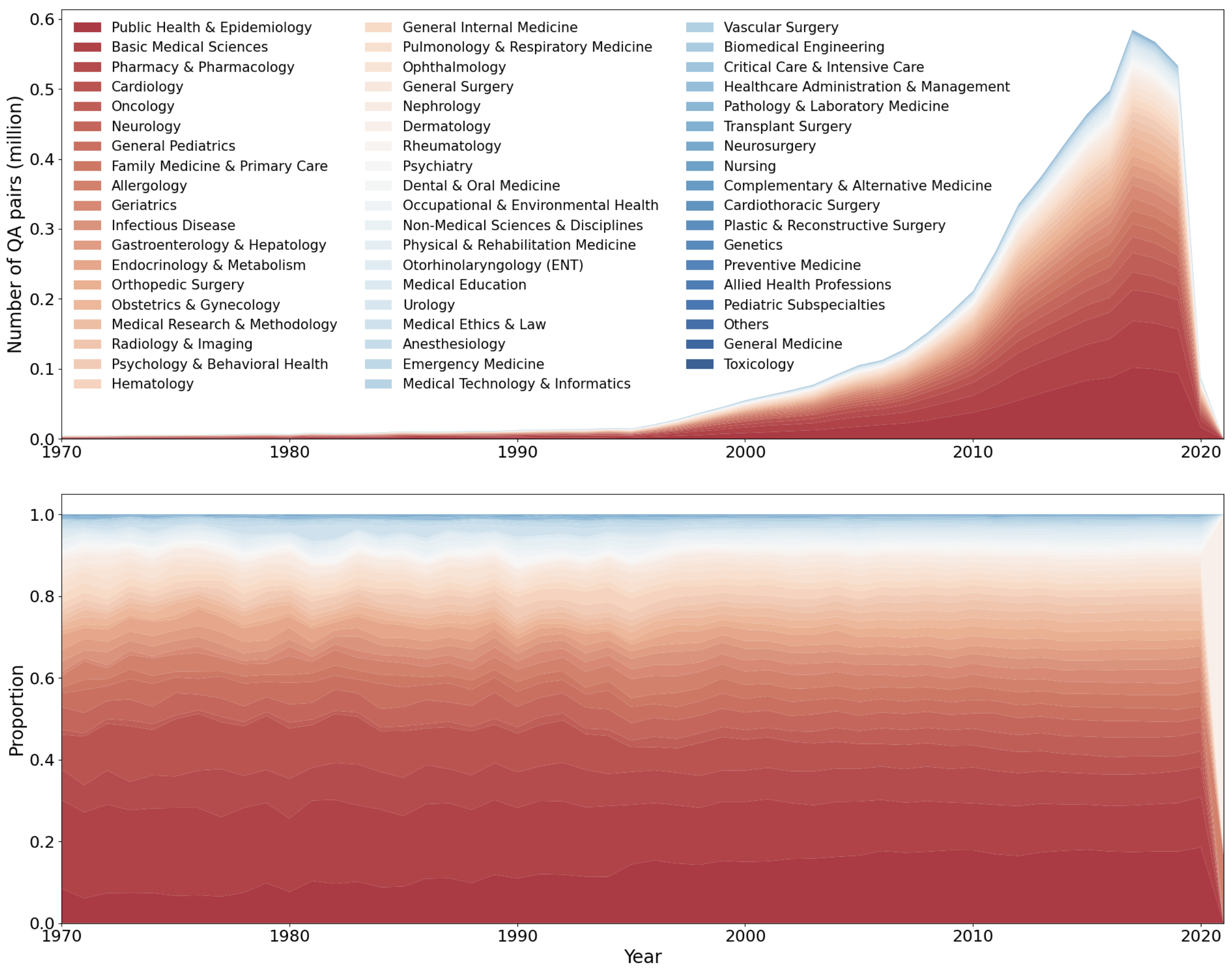}
    \caption{\textbf{A fine-grained version of the discipline and topics distribution of MIRIAD over time.} Here, all 56 categories of topics and disciplines for each question-answer (QA) pair are shown. The top panel shows absolute counts of QA pairs within a given year, the bottom row shows the relative proportion of fine-grained topics within a given year bucket.}
    \label{fig:sup_fig_fine_grained_disciplines}
\end{figure}
\begin{figure}[p]
    \centering
    \includegraphics[width=1\linewidth]{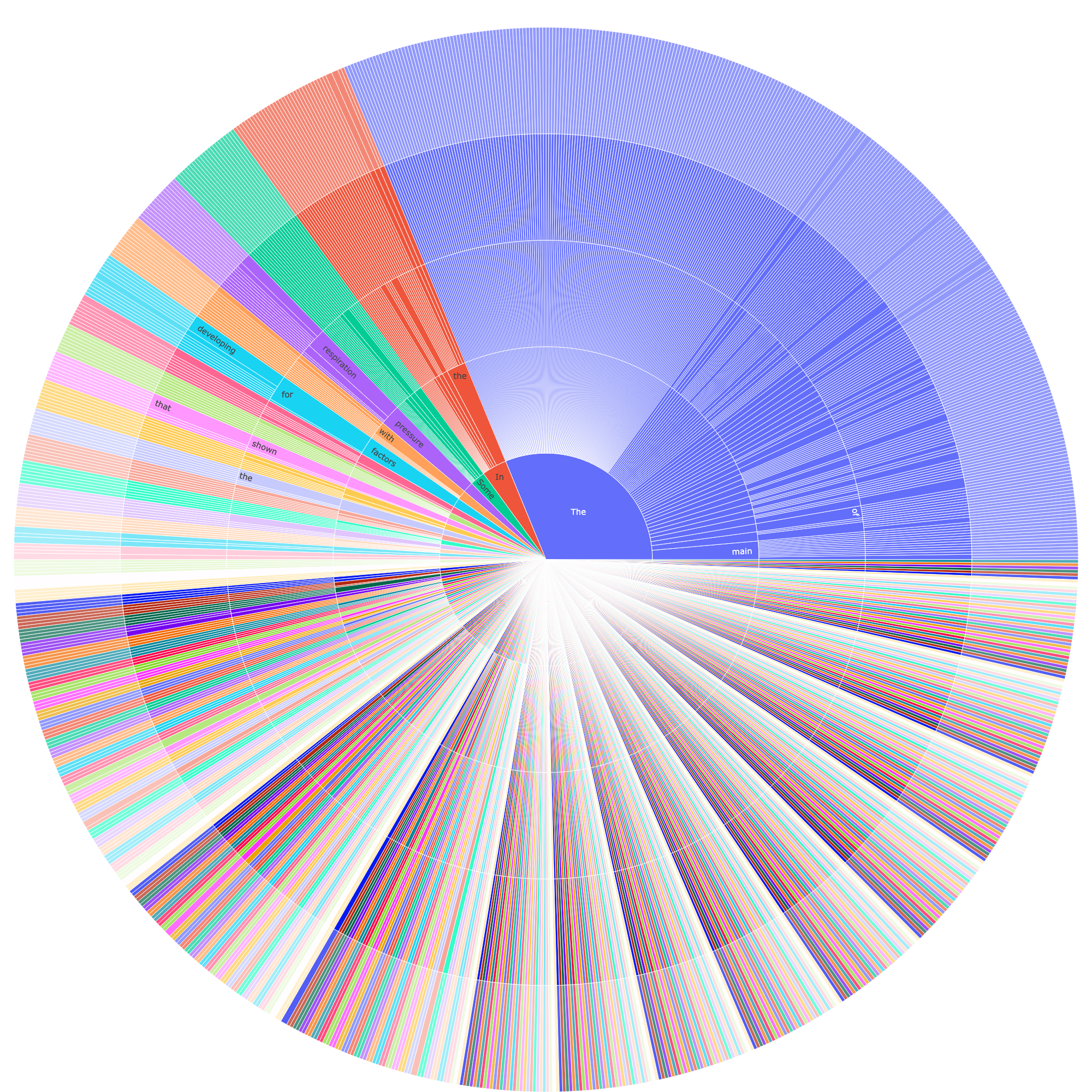}
    \caption{\textbf{Sunburst plot of MIRIAD answers.} The first five words of a small subset of 1000 MIRIAD answers are visualized hierarchically in a sunburst plot. Color indicates that the different answers share the same first word, e.g. “The” or “In”. This plot exemplifies the breath of answers contained in MIRIAD. When increasing the number of samples, the diversity further increases, making it more challenging to visualize and interpret.}
    \label{fig:sup_fig_sunburst}
\end{figure}
\clearpage

\section{Retrieval-augmented generation (RAG) for medical question answering: background, experimental details, additional results}
\label{Supp:rag-details}
\subsection{What is RAG?}

LLMs are trained on extensive corpora comprising books, Wikipedia articles, web content and scientific literature \cite{brown2020language, achiam2023gpt, touvron2023llama, grattafiori2024llama, wu2024pmc, OpenBioLLMs, moor2023med}. Through the pre-training and alignment process, LLMs acquire vast amounts of general, albeit static, knowledge, enabling them to perform remarkably well across a wide range of tasks, including open-domain question answering, summarization, and natural dialogue \cite{achiam2023gpt, grattafiori2024llama}. Their ability to generate fluent, coherent, and human-like responses has led to rapid adoption among the general public and across various industries. However, this static knowledge foundation presents critical limitations for high-stakes domains like medicine. First, LLMs often lack deep domain-specific expertise, particularly in specialized or rapidly evolving areas of clinical research. As a result, they frequently produce superficial responses when detailed, nuanced understanding is required. Second, updating their internal knowledge is computationally expensive and slow. Models with billions of parameters are often trained on trillions of diverse and high-quality tokens \cite{liu2024deepseek, penedo2024fineweb}, making routine updates impractical and slow, especially in fast-moving fields requiring up-to-date information. Third, they are prone to hallucinations, producing plausible-sounding yet factually incorrect responses, especially when confronted with rare or complex questions. Lastly, most LLMs lack mechanisms for source attribution, offering little transparency or evidence to support their claims. This absence of grounding undermines trust and makes it difficult for users to verify information. These limitations can be practically mitigated through Retrieval-Augmented Generation (RAG).

\subsection{How is RAG conventionally implemented?}
Retrieval-Augmented Generation (RAG) is an emerging hybrid architecture designed to address the limitations of generative language models. It consists of two main components: (i) a retrieval module, which identifies relevant documents from an external knowledge source, and (ii) a generation module, typically a transformer-based LLM, which synthesizes responses conditioned on the retrieved content. The retrieval module in RAG typically leverages dense vector representations to embed queries and identify relevant documents from large databases. Once retrieved, these documents are passed to the generative module, which will take care of context integration and generation. This design enables RAG models to produce fluent, context-aware text that is grounded in real-world and up-to-date information. By incorporating external retrieval at inference time, RAG helps mitigate common issues such as factual inaccuracies and hallucinations. It enhances the factual reliability of outputs and allows for greater transparency. RAG has been applied across a range of domains, including open-domain question answering, conversational assistants, and personalized recommendation systems. It is increasingly being explored in areas requiring reliable information support such as biomedicine.

\subsection{Prompts for RAG experiments}\label{supp:rag-prompts}
Here we provide the prompts we used for the RAG experiment. The results can be found in Section~\ref{sec:rag-results}. 

\begin{tcolorbox}[colback=gray!5!white, colframe=gray!89!black, title=System Prompt With RAG]
You are an expert medical student. You will be given a clinical knowledge question and a series of answer choices. You will be given examples of similar questions and their ideal answers or relevant passage text chunks.\\

You want to answer in strict JSON format, with 2 keys:
\begin{itemize}
    \item "choice": a letter (ie A, B, C, D, E) corresponding to the correct answer choice
    \item "answer": Where you explain your choice and reasoning
\end{itemize}

i.e. the output should be like: {"choice": (single letter), "answer": text for answer explanation}. No need to elaborate other than the json output.
\end{tcolorbox}

\begin{tcolorbox}[colback=gray!5!white, colframe=gray!89!black, title=User Prompt With RAG]
\# Similar Examples\\

$\{retrieved\_RAG\_samples\}$\\

\# Question\\

$\{benchmark\_MCQ\}$\\

\# Choices\\

$\{corresponding\_choices\}$

\end{tcolorbox}

\begin{tcolorbox}[colback=gray!5!white, colframe=gray!89!black, title=System Prompt Without RAG]
You are an expert medical student. You will be given a clinical knowledge question and a series of answer choices.\\

You want to answer in strict JSON format, with 2 keys:
\begin{itemize}
    \item "choice": a letter (ie A, B, C, D, E) corresponding to the correct answer choice
    \item "answer": Where you explain your choice and reasoning
\end{itemize}
\end{tcolorbox}

\begin{tcolorbox}[colback=gray!5!white, colframe=gray!89!black, title=User Prompt Without RAG]
\# Question\\

$\{benchmark\_MCQ\}$\\

\# Choices\\

$\{correspondin\_choices\}$

\end{tcolorbox}

\subsection{Additional RAG results with different n-tokens} \label{supp:rag-tokens}
In Table~\ref{tab:additional_rag_results_n_tokens}, we present the accuracy (with 95\% confidence intervals) of RAG-MIRIAD and RAG-Passage across increasing context window sizes (n-tokens $\in$ \{200, 600, 1000, 1400, 1800, 2200, 2600\}), evaluated on MedMCQA. Results are shown for combinations of two embedding models (BGE-Large, all-MiniLM) and two backbone LLMs (Mixtral-8x7B-instruct, Llama-3.1-8B-instruct). The No-RAG baseline remains constant across context lengths and is included once per configuration for clarity.

As the context window expands, the accuracies of all RAG variants generally increase, but RAG-MIRIAD demonstrates faster and more stable gains in performance with the same amount of tokens. This effect is most pronounced under the BGE-Large embedding configuration, where RAG-MIRIAD begins outperforming No-RAG at just 200 tokens. In contrast, RAG-Passage requires longer windows (often >600 tokens) to consistently surpass the baseline. These results reinforce the efficiency of structured QA-based retrieval: MIRIAD enables more effective knowledge integration under strict context budgets.

\begin{table}
\centering
\scriptsize
\centering
\resizebox{\textwidth}{!}{%
\begin{tabular}{llc|cc}
\toprule
\textbf{Embed Model} & \textbf{LLM} & \textbf{n-tokens} & \textbf{RAG-MIRIAD} & \textbf{RAG-Passage} \\
\midrule
\multirow{14}{*}{bge-large-en-v1.5} 
& \multirow{6}{*}{Mixtral-8x7B-Instruct}
& 200 & \textbf{55.13 [53.67, 56.61]} & 52.83 [51.35, 54.36] \\
& & 600 & \textbf{58.07 [56.63, 59.55]} & 54.46 [53.00, 55.96] \\
& & 1000 & \textbf{59.69 [58.26, 61.20]} & 55.92 [54.46, 57.45] \\
& & 1400 & \textbf{58.62 [57.18, 60.10]} & 56.71 [55.22, 58.21] \\
& & 1800 & \textbf{59.38 [57.90, 60.89]} & 57.28 [55.80, 58.81] \\
& & 2200 & \textbf{59.22 [57.76, 60.70]} & 57.26 [55.77, 58.79] \\
& & 2600 & \textbf{59.34 [57.88, 60.84]} & 57.54 [56.08, 59.07] \\
\cmidrule(l){3-5}
& &\multicolumn{1}{c}{No-RAG} & \multicolumn{1}{c}{53.48 [52.00, 55.01]} &\\

\cmidrule(l){2-5}

& \multirow{7}{*}{Llama-3.1-8B-Instruct}
& 200 & \textbf{56.85 [55.34, 58.38]} & 53.84 [52.38, 55.32] \\
& & 600 & \textbf{58.07 [56.63, 59.55]} & 55.39 [53.93, 56.90] \\
& & 1000 & \textbf{59.36 [57.88, 60.87]} & 55.94 [54.48, 57.47] \\
& & 1400 & \textbf{58.86 [57.40, 60.36]} & 56.49 [55.01, 57.97] \\
& & 1800 & \textbf{58.91 [57.45, 60.41]} & 56.47 [54.98, 57.95] \\
& & 2200 & \textbf{58.93 [57.47, 60.44]} & 56.49 [55.01, 57.97] \\
& & 2600 & \textbf{59.02 [57.57, 60.53]} & 56.78 [55.30, 58.28] \\
\cmidrule(l){3-5}
& &\multicolumn{1}{c}{No-RAG} & \multicolumn{1}{c}{56.08 [54.63, 57.61]} &\\

\midrule

\multirow{14}{*}{all-MiniLM-L6-v2} 
& \multirow{7}{*}{Mixtral-8x7B-Instruct}
& 200 & \textbf{54.58 [53.10, 56.08]} & 52.76 [51.28, 54.27] \\
& & 600 & \textbf{56.11 [54.67, 57.64]} & 54.05 [52.59, 55.58] \\
& & 1000 & \textbf{56.37 [54.89, 57.85]} & 54.91 [53.45, 56.4] \\
& & 1400 & \textbf{56.73 [55.22, 58.24]} & 55.01 [53.55, 56.51] \\
& & 1800 & \textbf{57.66 [56.20, 59.19]} & 54.98 [53.53, 56.49] \\
& & 2200 & \textbf{57.14 [55.65, 58.67]} & 54.94 [53.48, 56.44] \\
& & 2600 & \textbf{57.38 [55.89, 58.88]} & 55.08 [53.62, 56.56] \\
\cmidrule(l){3-5}
& &\multicolumn{1}{c}{No-RAG} & \multicolumn{1}{c}{53.48 [52.00, 55.01]} &\\

\cmidrule(l){2-5}

& \multirow{7}{*}{Llama-3.1-8B-Instruct}
& 200 & \textbf{54.98 [53.53, 56.49]} & 52.83 [51.35, 54.36] \\
& & 600 & \textbf{55.32 [53.86, 56.78]} & 53.12 [51.64, 54.65] \\
& & 1000 & \textbf{56.11 [54.67, 57.64]} & 53.07 [51.59, 54.58] \\
& & 1400 & \textbf{56.32 [54.86, 57.83]} & 53.88 [52.43, 55.37] \\
& & 1800 & \textbf{56.3 [54.84, 57.81]} & 53.55 [52.09, 55.06] \\
& & 2200 & \textbf{55.89 [54.43, 57.42]} & 53.62 [52.16, 55.13] \\
& & 2600 & \textbf{56.28 [54.82, 57.81]} & 54.46 [53.00, 55.96] \\
\cmidrule(l){3-5}
& &\multicolumn{1}{c}{No-RAG} & \multicolumn{1}{c}{56.08 [54.63, 57.61]} &\\
\bottomrule
\end{tabular}
}
\caption{\textbf{RAG results on the MedMCQA dataset across varying context budget (n-tokens).} Results are reported as accuracy (with 95\% confidence intervals). Across all evaluated configurations, retrieval using MIRIAD’s structured QA pairs (RAG-MIRIAD) consistently outperforms retrieval from unstructured text chunks (RAG-Passage). This further validates the effectiveness of utilizing MIRIAD with RAG.}
\label{tab:additional_rag_results_n_tokens}
\end{table}
\clearpage

\subsection{Validation of RAG findings on further benchmarks} \label{supp:validation-benchmarks}
To evaluate whether the benefits of structured retrieval extend beyond MedMCQA, we tested the same setup—using bge-large-en-v1.5 for embedding and a 1000-token retrieval window—on two additional medical benchmarks: MMLU-Med and MedQA-USMLE. As shown in the following table, RAG-MIRIAD continues to outperform RAG-Passage across both models and datasets. As shown in Table~\ref{tab:rag_comparison5.8M_w_CI_1000tokens_additional_benchmarks_mmlu_medqa_BGEonly}, on MMLU-Med, Mixtral-8x7B-Instruct showed a relative gain of 3.21\%, and Llama-3.1-8B-Instruct improved by 1.73\%. Similarly, on MedQA-USMLE, Mixtral improved by 2\%, and Llama by 0.7\%. These consistent gains indicate that MIRIAD’s structured format offers a retrieval advantage that generalizes across diverse medical QA tasks—not just due to domain relevance, but likely because of its condensed and answer-oriented information layout, which better supports knowledge integration during inference.

\begin{table}[h]
    \centering
    \resizebox{\textwidth}{!}{%
    \begin{tabular}{ccccc}
    \toprule
     \multirow{1}{*}{\textbf{Benchmark}} & \multirow{1}{*}{\textbf{Backbone LLM}} &  \textbf{\textit{RAG-MIRIAD}} & \textbf{\textit{RAG-Passage}} & \textbf{\textit{No-RAG}} \\
    \midrule
     \multirow{2}{*}{MMLU-Med} & Mixtral-8x7B-Instruct & \textbf{76.77 [74.2, 79.25]} & 74.38 [71.9, 76.95] & 76.03 [73.55, 78.6]\\
        & Llama-3.1-8B-Instruct & \textbf{75.02 [72.45, 77.59]} & 73.74 [71.17, 76.31] & 72.36 [69.61, 74.93]\\
    \midrule 
         \multirow{2}{*}{MedQA-USMLE} & Mixtral-8x7B-Instruct & \textbf{60.02 [57.19, 62.69]} & 58.84 [56.01, 61.51] & 57.82 [55.06, 60.41]\\
        & Llama-3.1-8B-Instruct & \textbf{61.27 [58.52, 64.02]} & 60.8 [58.05, 63.55] & 61.12 [58.44, 63.86]\\
    \bottomrule
    \end{tabular}
    }
    \caption{\textbf{Accuracy (with 95\% confidence intervals) of different RAG configurations on MMLU-Med and MedQA-US benchmarks.} Results compare RAG-MIRIAD, RAG-Passage, and No-RAG settings across two backbone LLMs (Mixtral-8x7B-Instruct and Llama-3.1-8B-Instruct), using BGE-Large embeddings and a 1000-token context budget.}
    \label{tab:rag_comparison5.8M_w_CI_1000tokens_additional_benchmarks_mmlu_medqa_BGEonly}
\end{table}
\clearpage

\section{Distribution of categories of MedMCQA (dev) problems} \label{supp:medmcqa-cat}
\begin{figure}[H]
    \centering
    \includegraphics[width=1\linewidth]{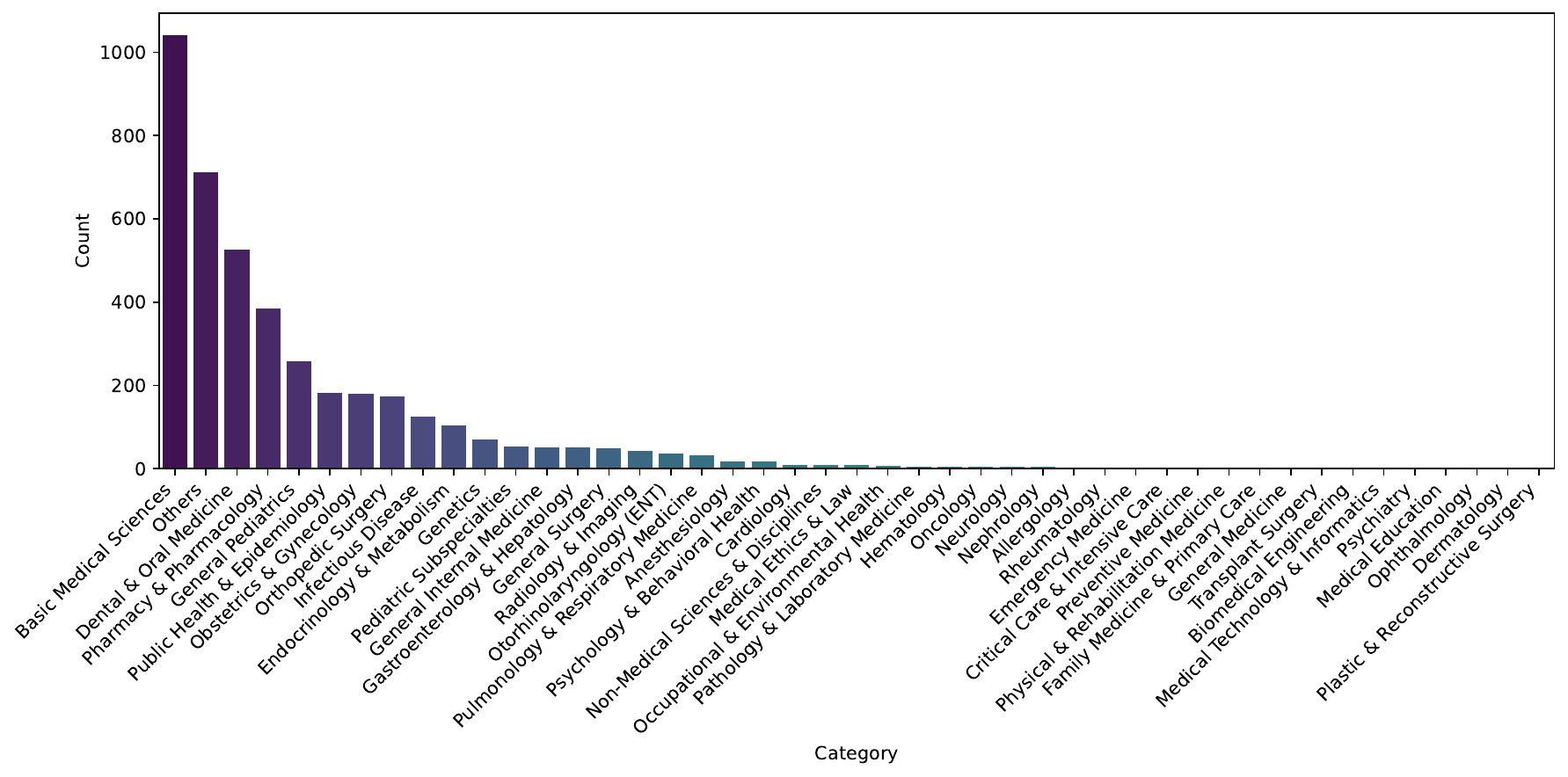}
    \caption{\textbf{Distribution of disciplines in MedMCQA dev set}}
    \label{fig:sup_fig_medmcqa_distr}
\end{figure}

Figure~\ref{fig:sup_fig_medmcqa_distr} illustrates the distribution of medical disciplines across the dev set of the MedMCQA dataset. Each question was categorized based on its domain using automatic annotation with the Llama-3-8B-Instruct model, followed by manual verification. The categories reflect a diverse range of medical specialties, including but not limited to cardiology, oncology, pharmacology, and genetics. This distribution provides insight into the domain coverage of MedMCQA, which is essential for discipline-specific impact analysis of RAG on MedMCQA.

\section{Additional details in medical discipline categorization}
To systematically assign a discipline-level category to each sample in MIRIAD, we employed the Llama-3.1-8b-Instruct model using a structured zero-shot prompting scheme. The goal was to categorize each sample into a fine-grained topic reflecting its underlying medical or biomedical domain. The prompt explicitly instructed the model to select the most appropriate topic from a predefined list of 50+ categories, based on the semantics of the question and its corresponding answer. To use the prompts, simply replace phrases in curly brackets with the actual content.
\begin{tcolorbox}[colback=gray!5!white, colframe=gray!89!black, title=MIRIAD Categorization Prompt]
\textbf{System Prompt}\\

You're a helpful medical assistant to help categorize the following question. Possible categorizations include:\\

Condition/Disease/Treatment/Symptom/Cause/Risk Factors/Prevention/Diagnosis/Prognosis/Pharmacology/Anatomy/Physiology/Biochemistry/Pathophysiology/Epidemiology/\\
Surgical Procedures/Nutrition and Diet/Genetics and Genomics/Pediatrics/Geriatrics/Psychology/\\
Psychiatry/Obstetrics and Gynecology/Dentistry/Immunology/Virology/Environmental and Occupational Health/Pharmacy and Drug Dispensation/Rehabilitation/Microbiology/Endocrinology/Neurology/Radiology/Oncology/Cardiology/Gastroenterology/Dermatology/Nephrology/Ophthalmology/\\
Orthopedics/Hematology/Rheumatology/Pulmonology/Urology/Otorhinolaryngology (ENT)/Veterinary Medicine/Addiction Medicine/Chiropractic Medicine/Palliative Care/Bioinformatics/Transplantation/Toxicology/Parasitology/Stem Cell Biology/Podiatry/Hepatology/Sports Medicine/Family Medicine/Sleep Medicine/Critical Care Medicine/Medical Ethics/Forensic Medicine/Infectious Diseases/Emergency Medicine\\

Please select no more than 4 reliable categories (it's possible to only select one), at least one of them has to be specialty. The chosen category has to be closely relevant to the question. Output in the following format. If there are multiple categories, separate them with comma. Please follow strictly the format below, no need to elaborate:\\

Category: xxx, xxx, xxx\\

\textbf{User Prompt}\\

Question: $\{MIRIAD\_Question\}$

\end{tcolorbox}

\subsection{Final stabilized medical disciplines}
\begin{figure}[H]
    \centering
    \includegraphics[width=1\linewidth]{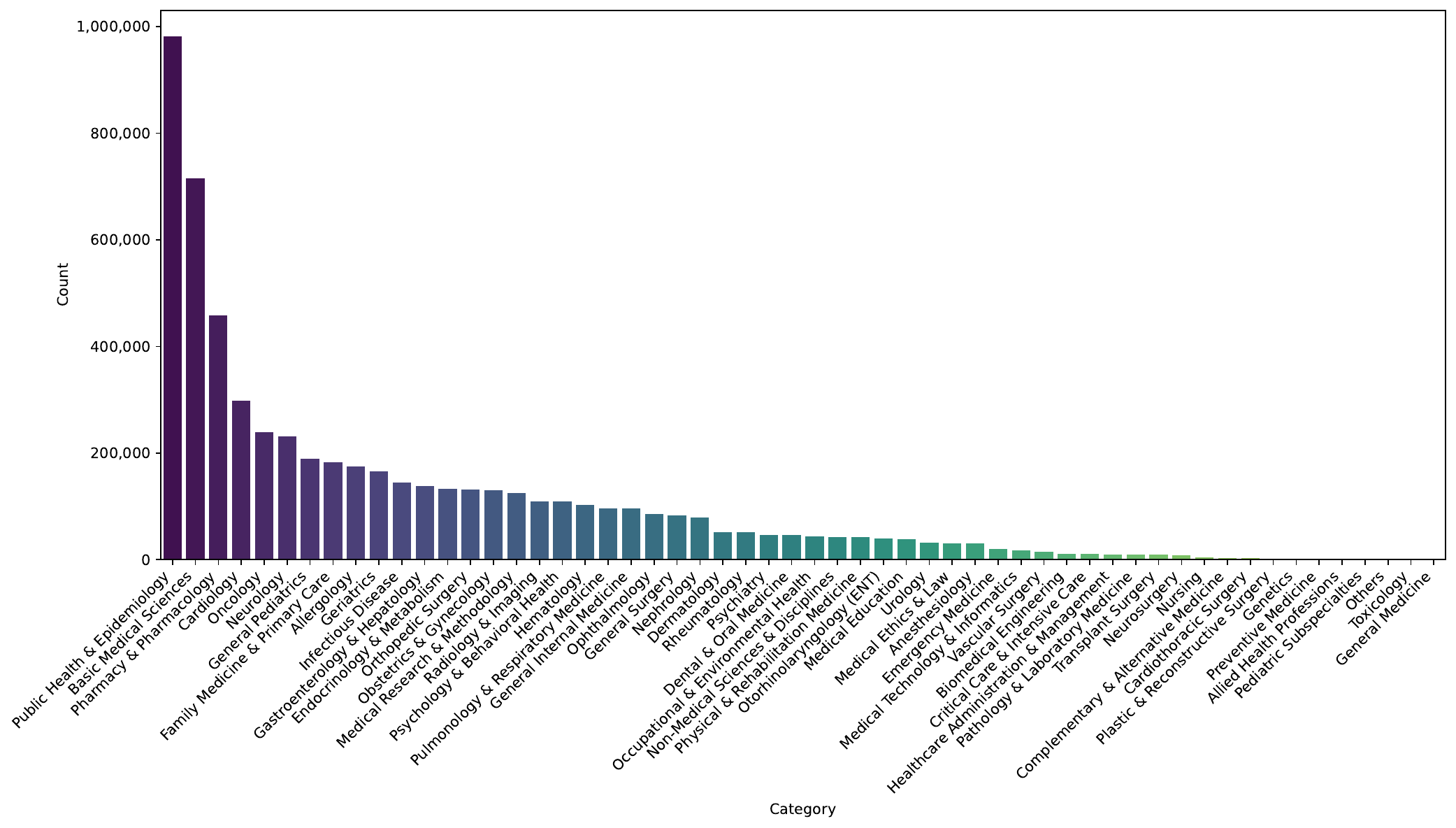}
    \caption{\textbf{Distribution of disciplines in MIRIAD}}
    \label{fig:sup_fig_miriad_distr}
\end{figure}

The raw Llama3 outputs were then mapped into a set of 56 stabilized and unified topic labels to ensure consistency across samples and reduce label fragmentation. This mapping process involved grouping similar categories and resolving overlaps or ambiguities under the guidance of medical expertise. The final distribution of 56 disciplines is visualized in Fig.~\ref{fig:sup_fig_miriad_distr}.\\

\section{Hallucination detection} \label{supp:hallucination_detection}
Here we provide the prompt templates used for the hallucination detection experiments described in Section~\ref{sec:medhallu-results} and \ref{sec:exp}. The prompts were designed to closely follow the structure proposed in the original MedHallu benchmark\cite{pandit2025medhallu}, ensuring systematic consistency with their evaluation protocol. Each prompt presents a question-answer pair, with or without additional retrieved evidence from MIRIAD, and instructs the model to classify the answer as either factual or hallucinated.

\begin{tcolorbox}[colback=gray!5!white, colframe=gray!89!black, title=System Prompt for Hallucination Detection - With MIRIAD Knowledge]
You are an AI assistant with extensive knowledge in the field of medicine. I want you to act as an answer judge. Given a question and an answer, your objective is to determine if the provided answer contains non-factual or hallucinated information. \\

You SHOULD give your judgement based on the following hallucination types and world knowledge.\\

You are trying to determine if there is a factual contradiction between the answer and the world knowledge.\\

Some information in the answer might be fabricated. Answer `0' if the answer is factual and `1' if the answer is hallucinated.\\

You can categorize the answer as hallucinated if it falls in either of the category:\\

Type: Misinterpretation of \#Question\#: These are hallucinated answers that misunderstand the question, leading to an off-topic or irrelevant response.\\

Type: Incomplete Information: These are hallucinated answers that Point out what is not true without providing correct information.\\

Type: Mechanism and Pathway Misattribution - These are hallucinated answers that falsely attribute biological mechanisms, molecular pathways, or disease processes that contradicts established medical knowledge.\\

Type: Methodological and Evidence Fabrication - Inventing false research methods, statistical data, or specific clinical outcomes.\\

Do not return anything else, just the answer.\\

Return just an integer value, `0' if the answer is factual and `1' if the answer is hallucinated. No letter or word, just the integer value.\\
\end{tcolorbox}

\begin{tcolorbox}[colback=gray!5!white, colframe=gray!89!black, title=User System Prompt for Hallucination Detection - With MIRIAD Knowledge]

World Knowledge: $\{external\_knowledge, i.e.\ retrieved\_MIRIAD\_QA\}$\\

Question: $\{benchmark\_question\}$\\

Answer: $\{paired\_answer\}$\\

Return just an integer value, `0' if the answer is factual and `1' if the answer is hallucinated. No letter or word, just the integer value.\\

Your Judgement:
\end{tcolorbox}

\begin{tcolorbox}[colback=gray!5!white, colframe=gray!89!black, title=System Prompt for Hallucination Detection - Without Knowledge]
You are an AI assistant with extensive knowledge in the field of medicine. I want you to act as an answer judge. Given a question and an answer, your objective is to determine if the provided answer contains non-factual or hallucinated information. \\

You SHOULD give your judgement based on the following hallucination types.\\

You are trying to determine if there is a factual contradiction between the answer and the world knowledge.\\

Some information in the answer might be fabricated. Answer `0' if the answer is factual and `1' if the answer is hallucinated.\\

You can categorize the answer as hallucinated if it falls in either of the category:\\

Type: Misinterpretation of \#Question\#: These are hallucinated answers that misunderstand the question, leading to an off-topic or irrelevant response.\\

Type: Incomplete Information: These are hallucinated answers that Point out what is not true without providing correct information.\\

Type: Mechanism and Pathway Misattribution - These are hallucinated answers that falsely attribute biological mechanisms, molecular pathways, or disease processes that contradicts established medical knowledge.\\

Type: Methodological and Evidence Fabrication - Inventing false research methods, statistical data, or specific clinical outcomes.\\

Do not return anything else, just the answer.\\

Return just an integer value, `0' if the answer is factual and `1' if the answer is hallucinated. No letter or word, just the integer value.\\
\end{tcolorbox}

\begin{tcolorbox}[colback=gray!5!white, colframe=gray!89!black, title=User System Prompt for Hallucination Detection - Without Knowledge]

Question: $\{benchmark\_question\}$\\

Answer: $\{paired\_answer\}$\\

Return just an integer value, `0' if the answer is factual and `1' if the answer is hallucinated. No letter or word, just the integer value.\\

Your Judgement:
\end{tcolorbox}

\section{Additional details in RAG individual sample contribution analysis} \label{supp:add-details-rag}

\begin{figure}[H]
    \centering
    \includegraphics[width=1\linewidth]{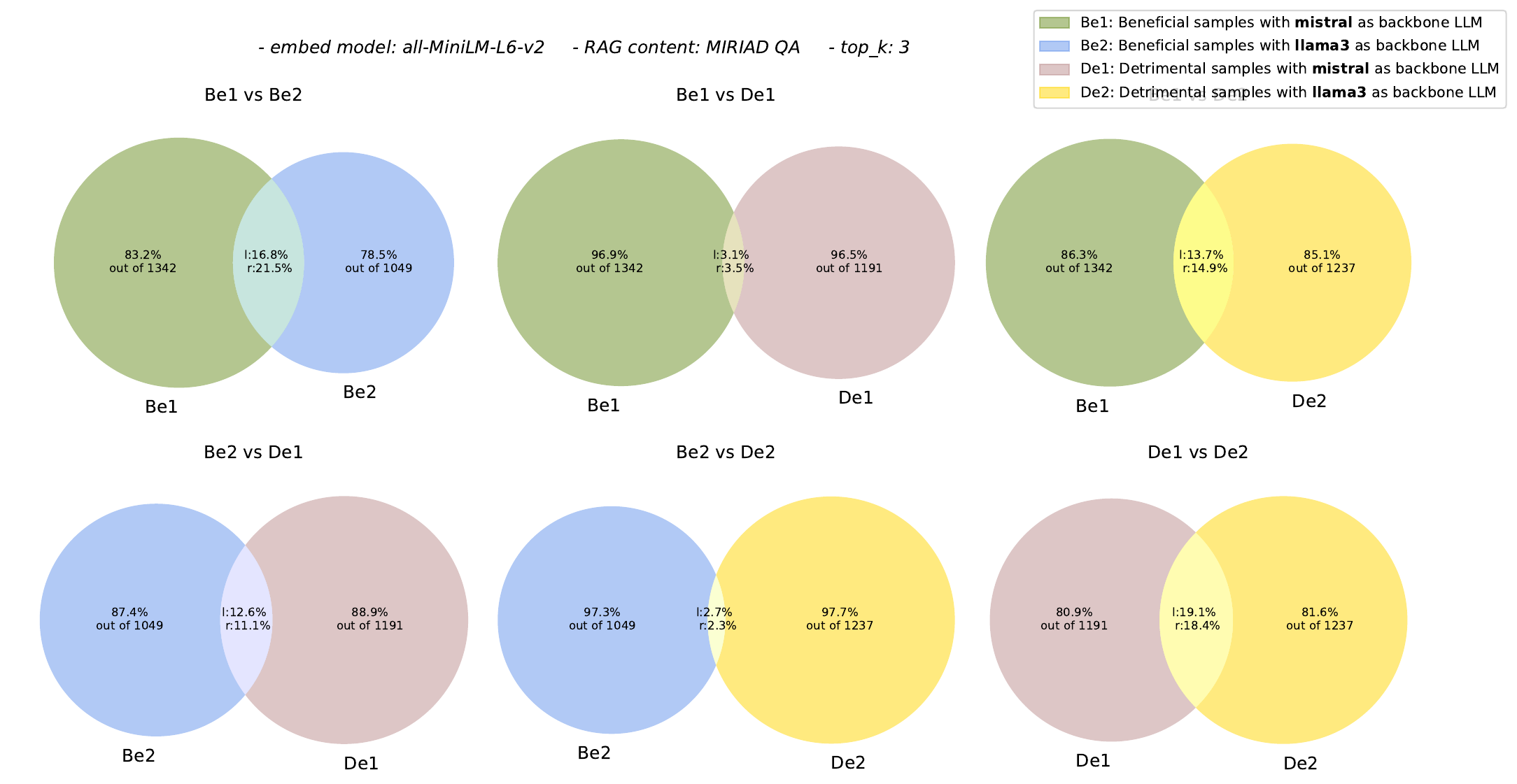}
    \caption{\textbf{Venn diagrams of beneficial and detrimental MIRAD samples for MedMCQA benchmark.} With fixed embedding model all-MiniLM, top-k as 3, and different backbone LLMs.}
    \label{fig:allminilm_qa_different_backbone_top3}
\end{figure}

\begin{figure}
    \centering
    \includegraphics[width=1\linewidth]{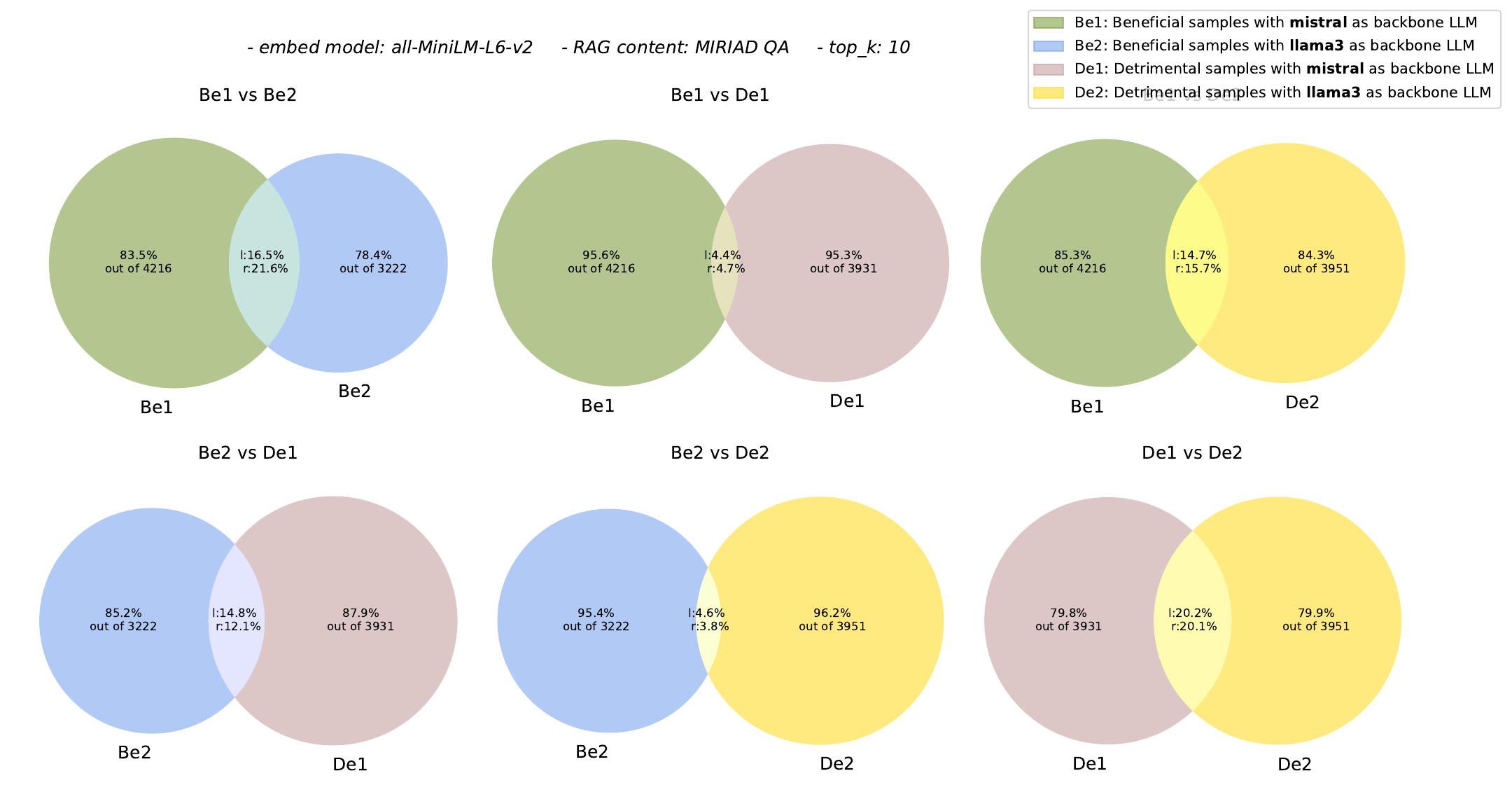}
    \caption{\textbf{Venn diagrams of beneficial and detrimental MIRAD samples for MedMCQA benchmark.} With fixed embedding model all-MiniLM, top-k as 10, and different backbone LLMs.}
    \label{fig:allminilm_qa_different_backbone_top10}
\end{figure}

\begin{figure}
    \centering
    \includegraphics[width=1\linewidth]{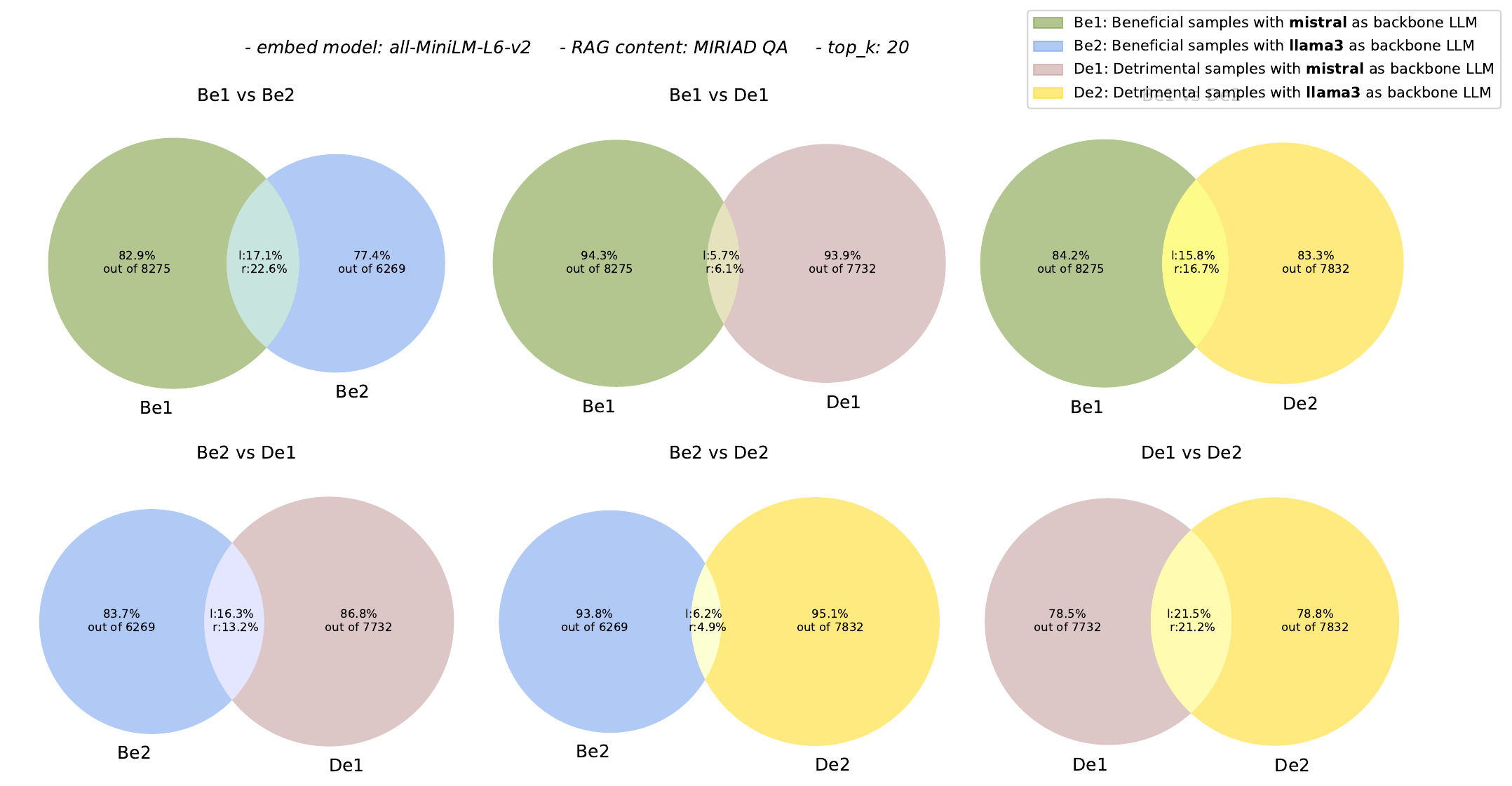}
    \caption{\textbf{Venn diagrams of beneficial and detrimental MIRAD samples for MedMCQA benchmark.} With fixed embedding model all-MiniLM, top-k as 20, and different backbone LLMs.}
    \label{fig:allminilm_qa_different_backbone_top20}
\end{figure}

\begin{figure}
    \centering
    \includegraphics[width=1\linewidth]{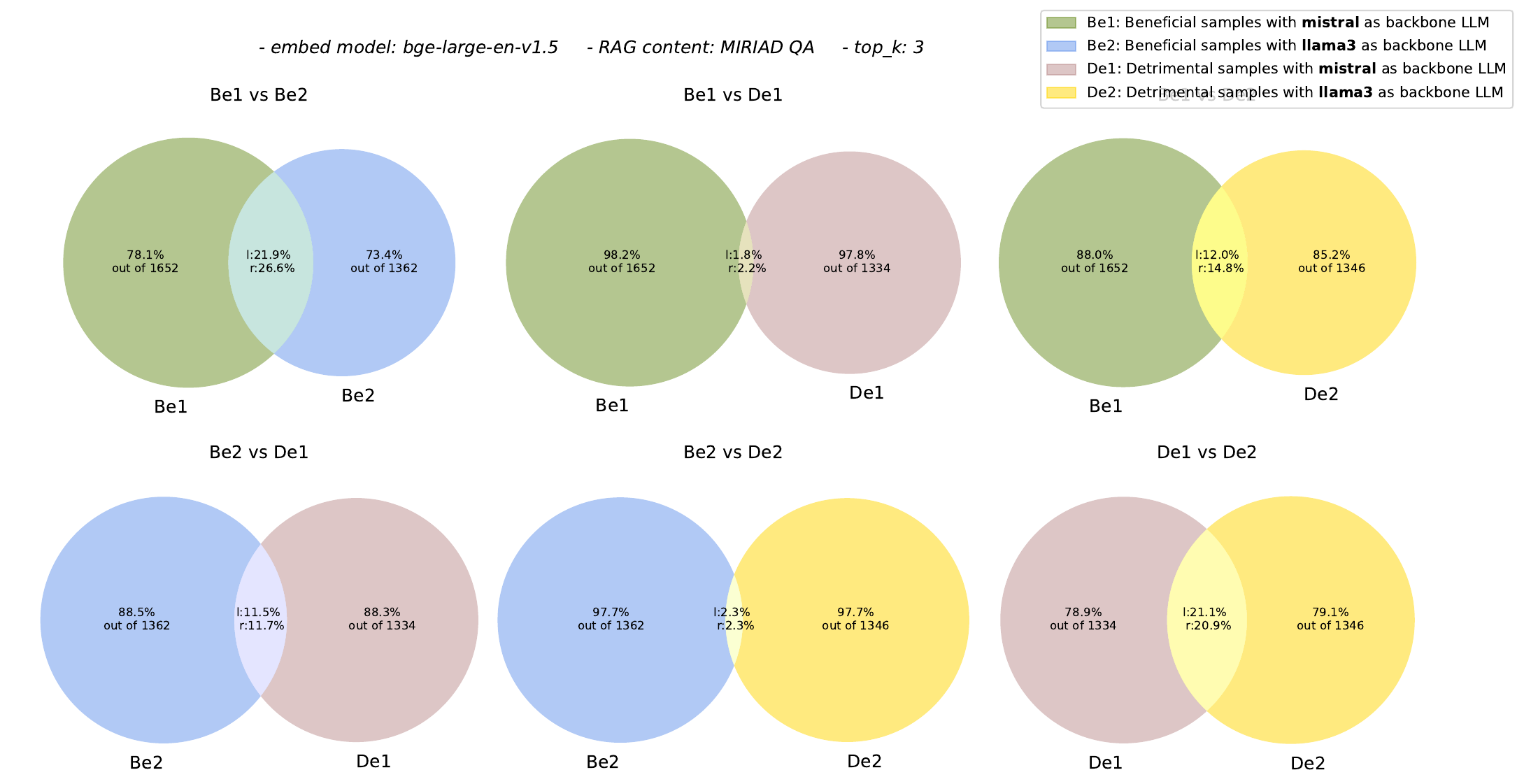}
    \caption{\textbf{Venn diagrams of beneficial and detrimental MIRAD samples for MedMCQA benchmark.} With fixed embedding model BGE-Large, top-k as 3, and different backbone LLMs.}
    \label{fig:bgelarge_qa_different_backbone_top3}
\end{figure}

\begin{figure}
    \centering
    \includegraphics[width=1\linewidth]{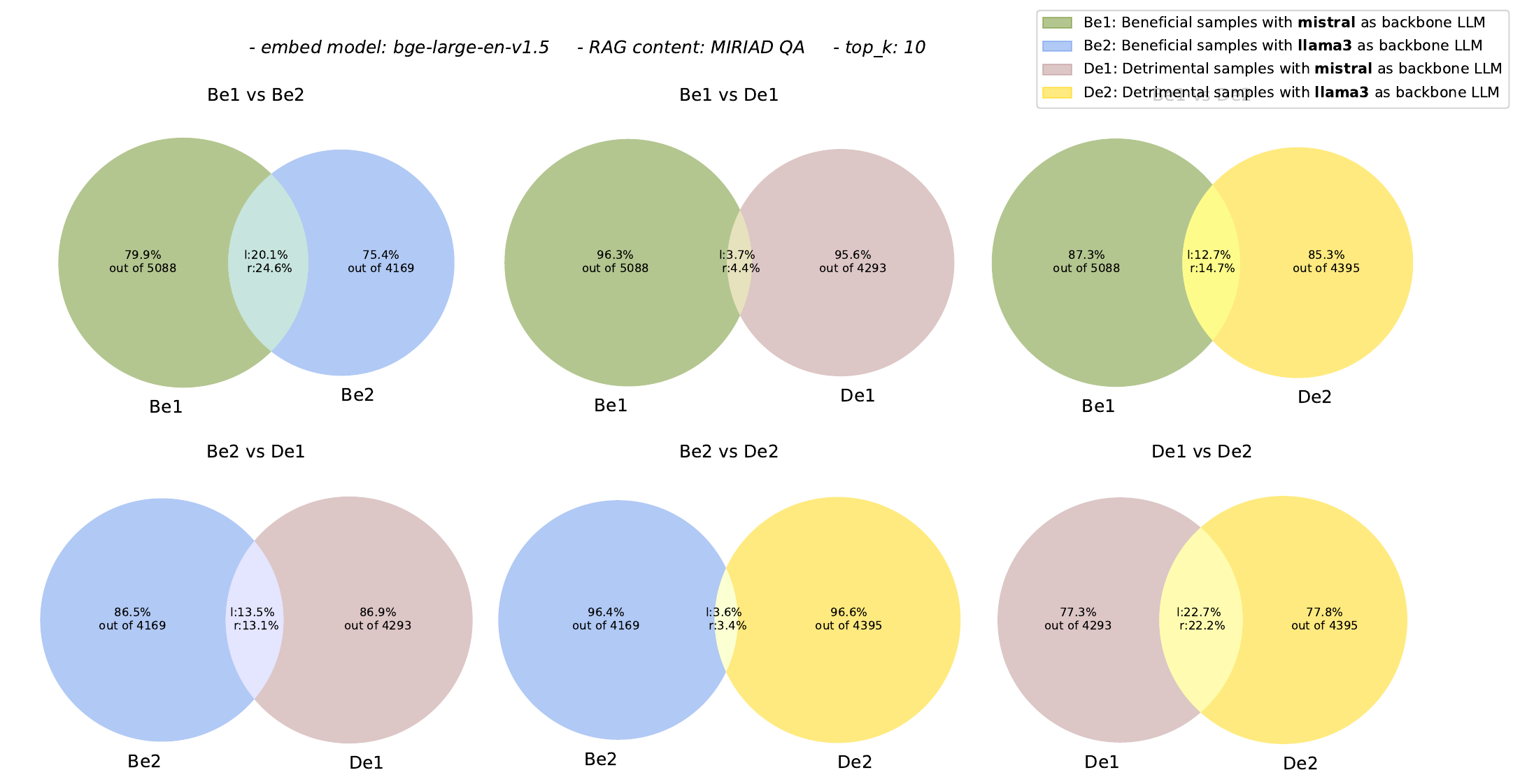}
    \caption{\textbf{Venn diagrams of beneficial and detrimental MIRAD samples for MedMCQA benchmark.} With fixed embedding model BGE-Large, top-k as 10, and different backbone LLMs.}
    \label{fig:bgelarge_qa_different_backbone_top10}
\end{figure}

\begin{figure}
    \centering
    \includegraphics[width=1\linewidth]{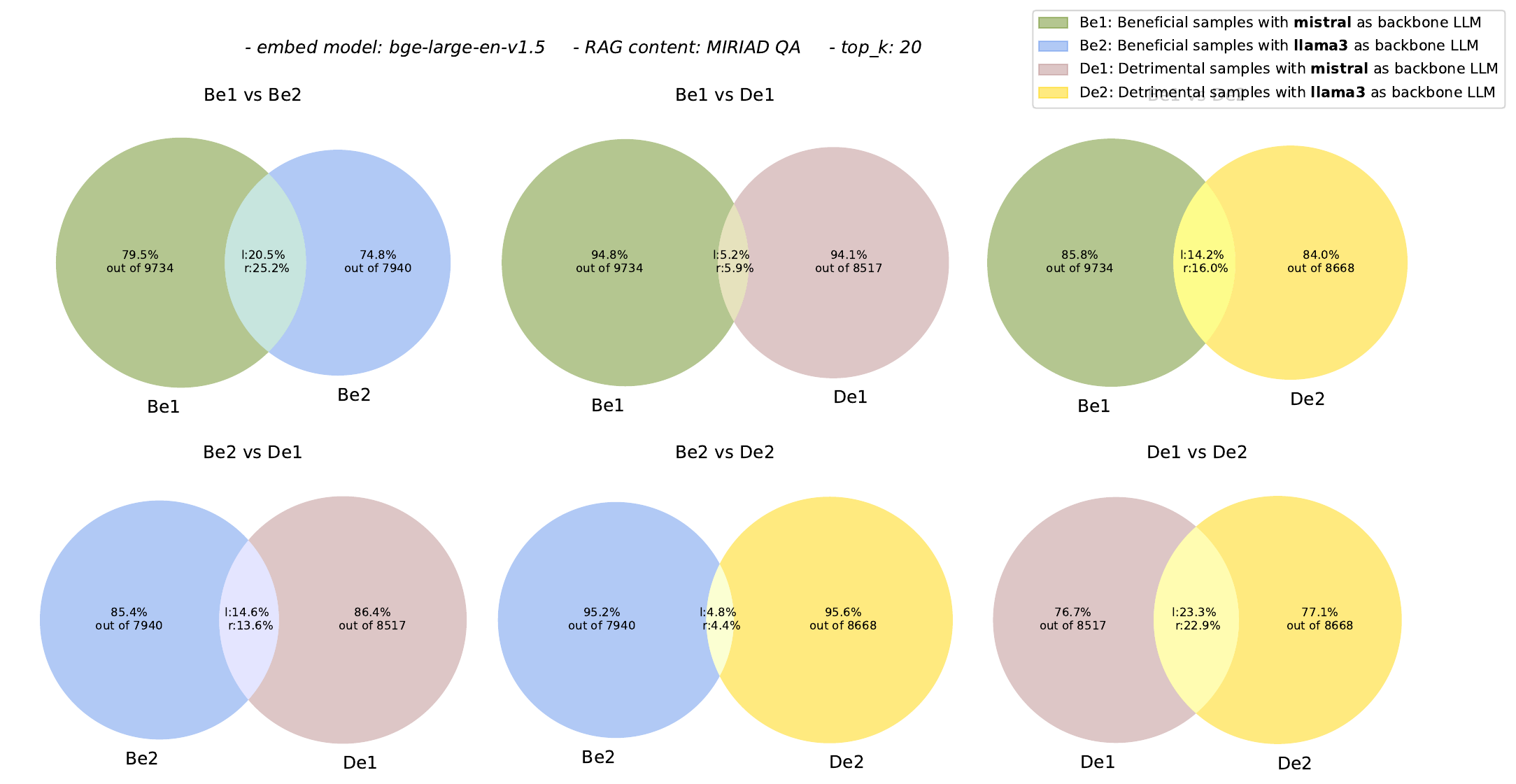}
    \caption{\textbf{Venn diagrams of beneficial and detrimental MIRAD samples for MedMCQA benchmark.} With fixed embedding model BGE-Large, top-k as 20, and different backbone LLMs.}
    \label{fig:bgelarge_qa_different_backbone_top20}
\end{figure}

\begin{figure}
    \centering
    \includegraphics[width=1\linewidth]{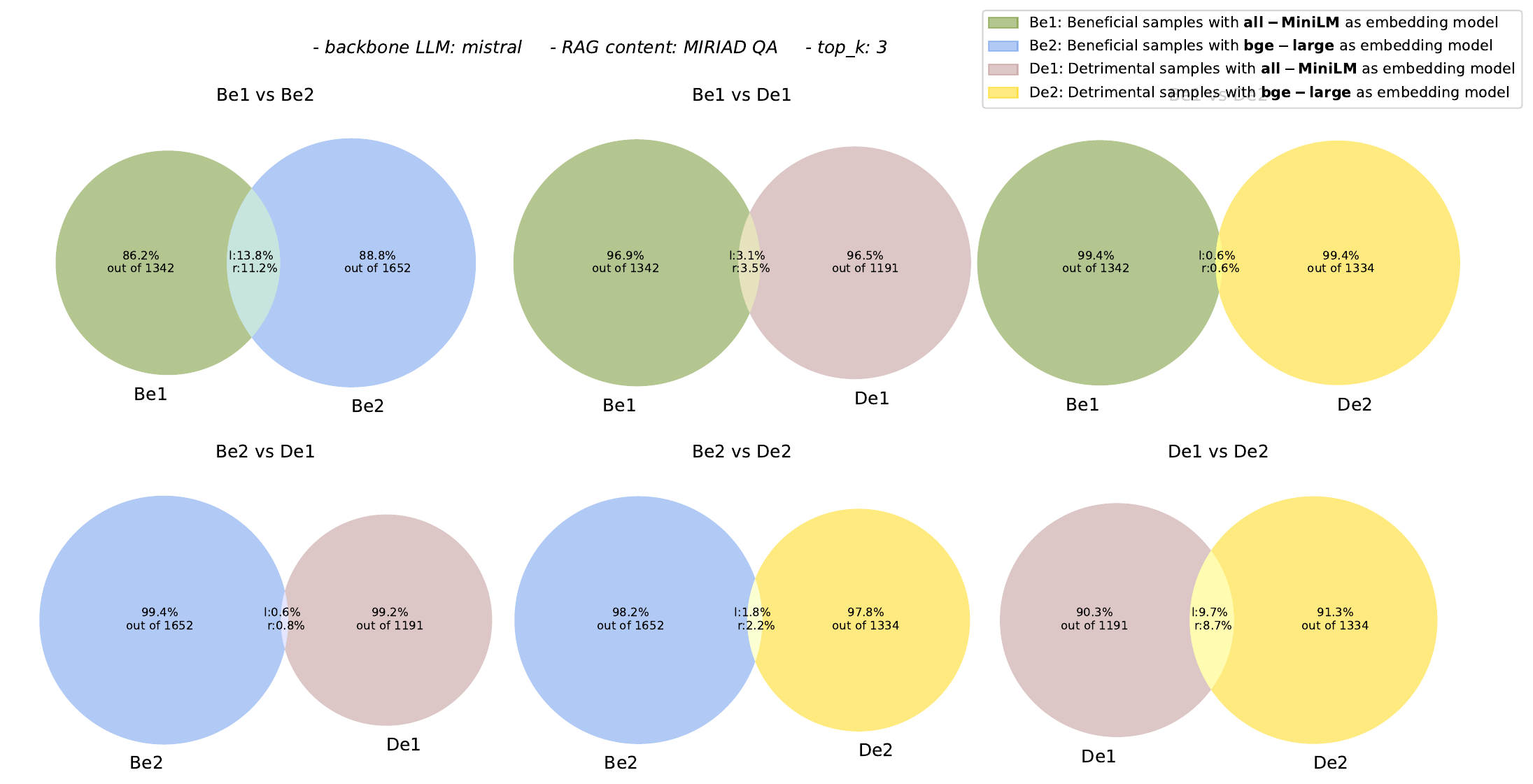}
    \caption{\textbf{Venn diagrams of beneficial and detrimental MIRAD samples for MedMCQA benchmark.} With fixed backbone LLM Mixtral-8x7B-Instruct, top-k as 3, and different embedding models.}
    \label{fig:mistral_qa_different_embed_top3}
\end{figure}

\begin{figure}
    \centering
    \includegraphics[width=1\linewidth]{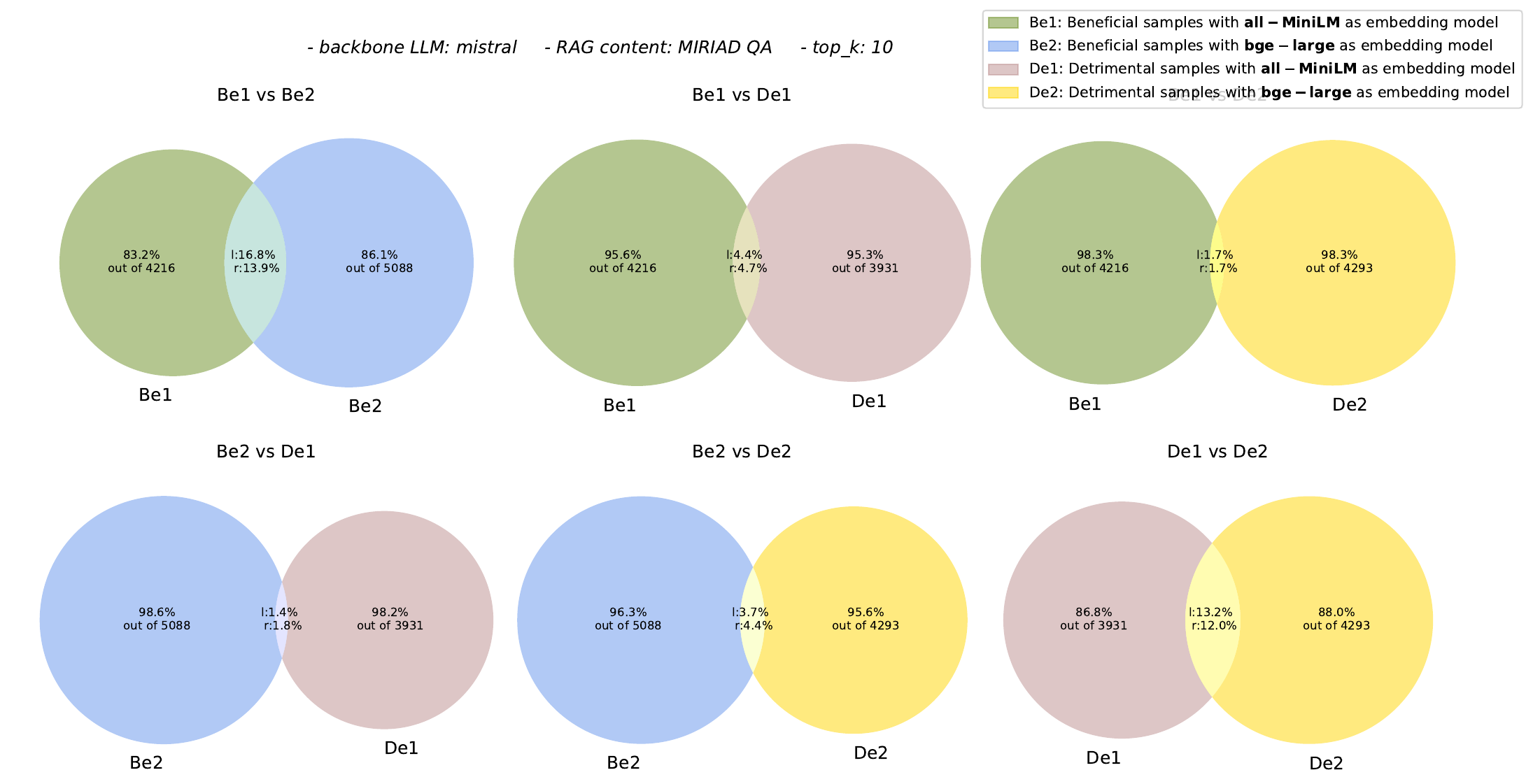}
    \caption{\textbf{Venn diagrams of beneficial and detrimental MIRAD samples for MedMCQA benchmark.} With fixed backbone LLM Mixtral-8x7B-Instruct, top-k as 10, and different embedding models.}
    \label{fig:mistral_qa_different_embed_top10}
\end{figure}

\begin{figure}
    \centering
    \includegraphics[width=1\linewidth]{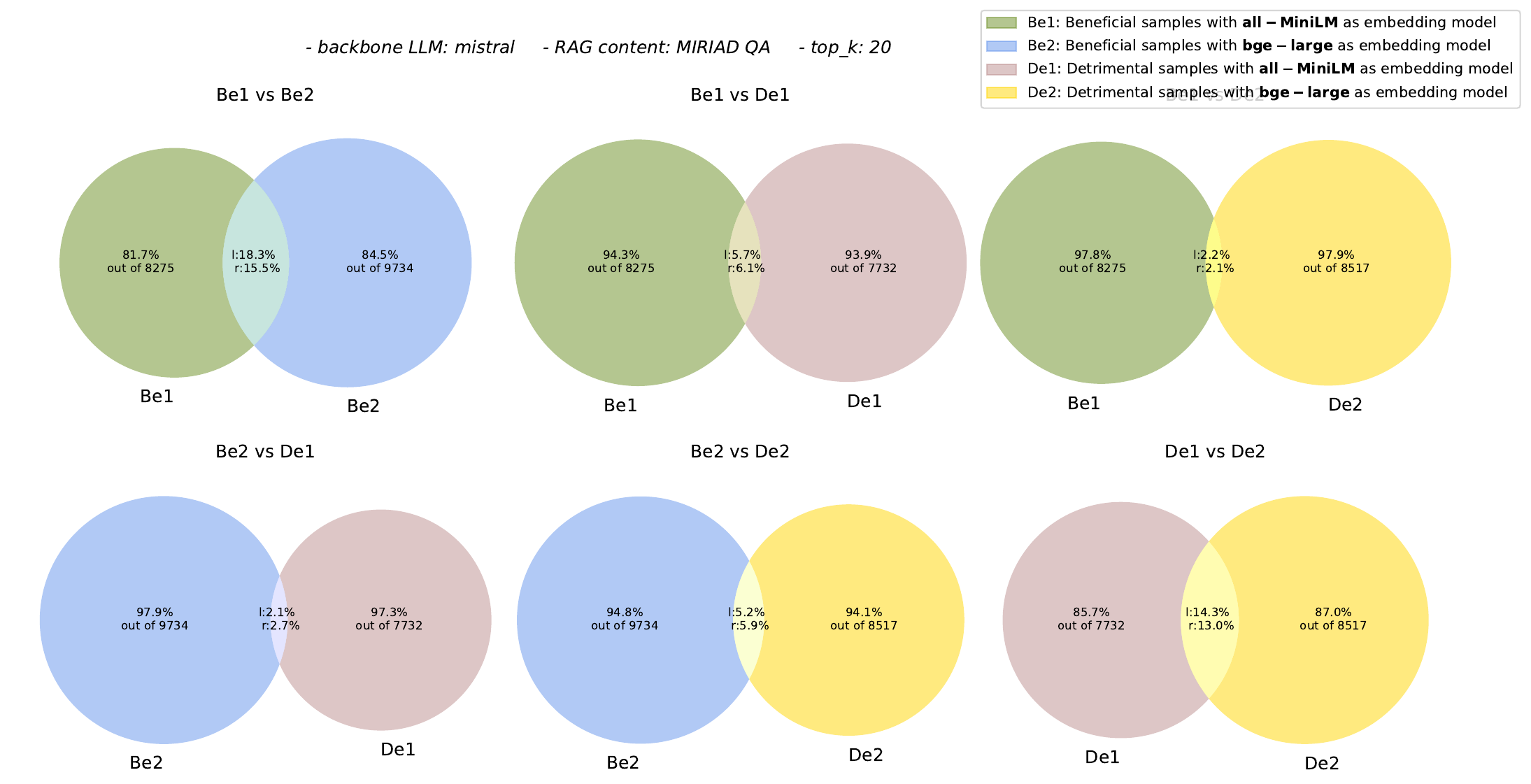}
    \caption{\textbf{Venn diagrams of beneficial and detrimental MIRAD samples for MedMCQA benchmark.} With fixed backbone LLM Mixtral-8x7B-Instruct, top-k as 20, and different embedding models.}
    \label{fig:mistral_qa_different_embed_top20}
\end{figure}

\begin{figure}
    \centering
    \includegraphics[width=1\linewidth]{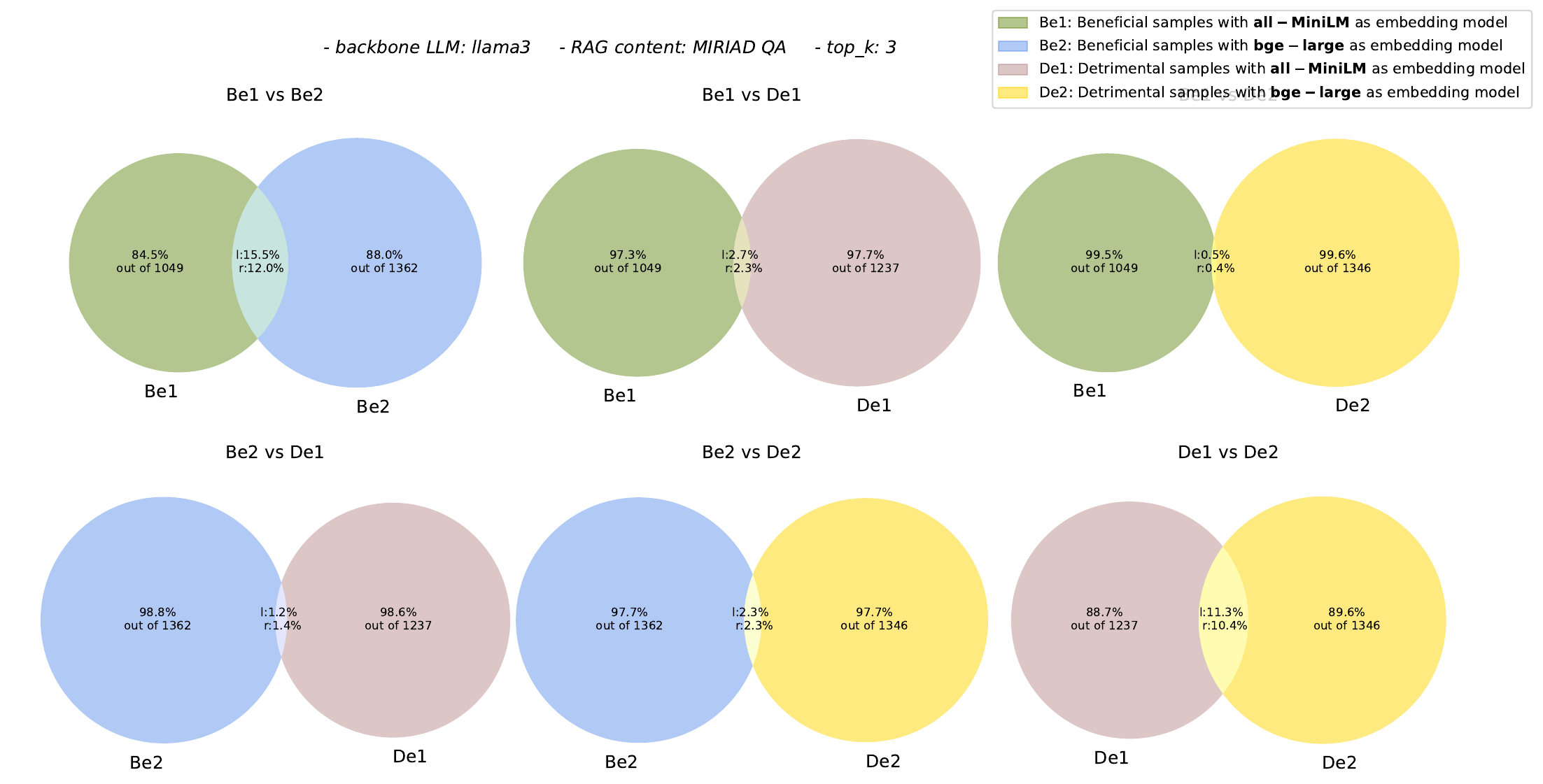}
    \caption{\textbf{Venn diagrams of beneficial and detrimental MIRAD samples for MedMCQA benchmark.} With fixed backbone LLM Llama-3.1-8B-instruct, top-k as 3, and different backbone embedding models.}
    \label{fig:llama3_qa_different_embed_top3}
\end{figure}

\begin{figure}
    \centering
    \includegraphics[width=1\linewidth]{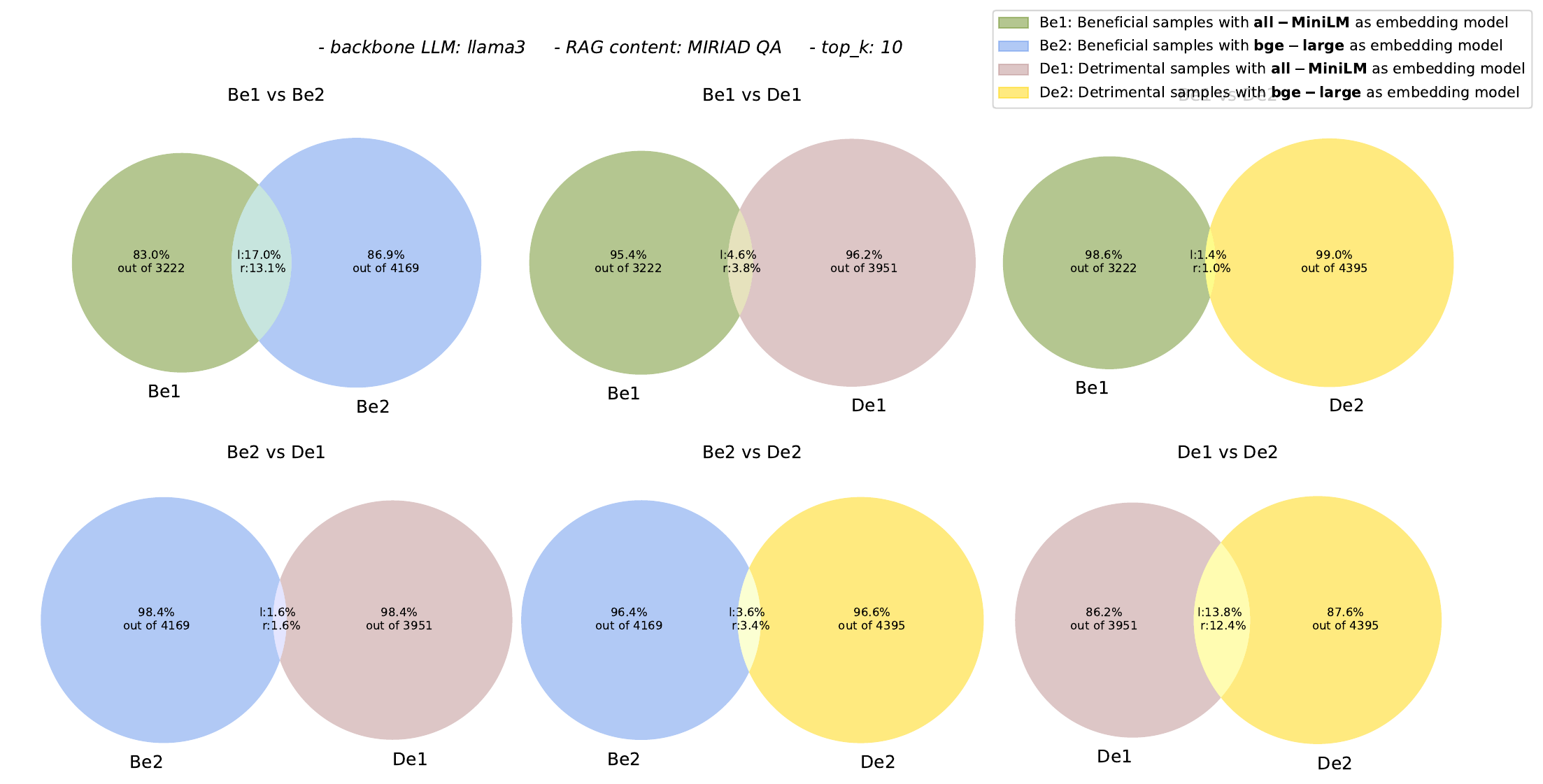}
    \caption{\textbf{Venn diagrams of beneficial and detrimental MIRAD samples for MedMCQA benchmark.} With fixed backbone LLM Llama-3.1-8B-instruct, top-k as 10, and different backbone embedding models.}
    \label{fig:llama3_qa_different_embed_top10}
\end{figure}

\begin{figure}
    \centering
    \includegraphics[width=1\linewidth]{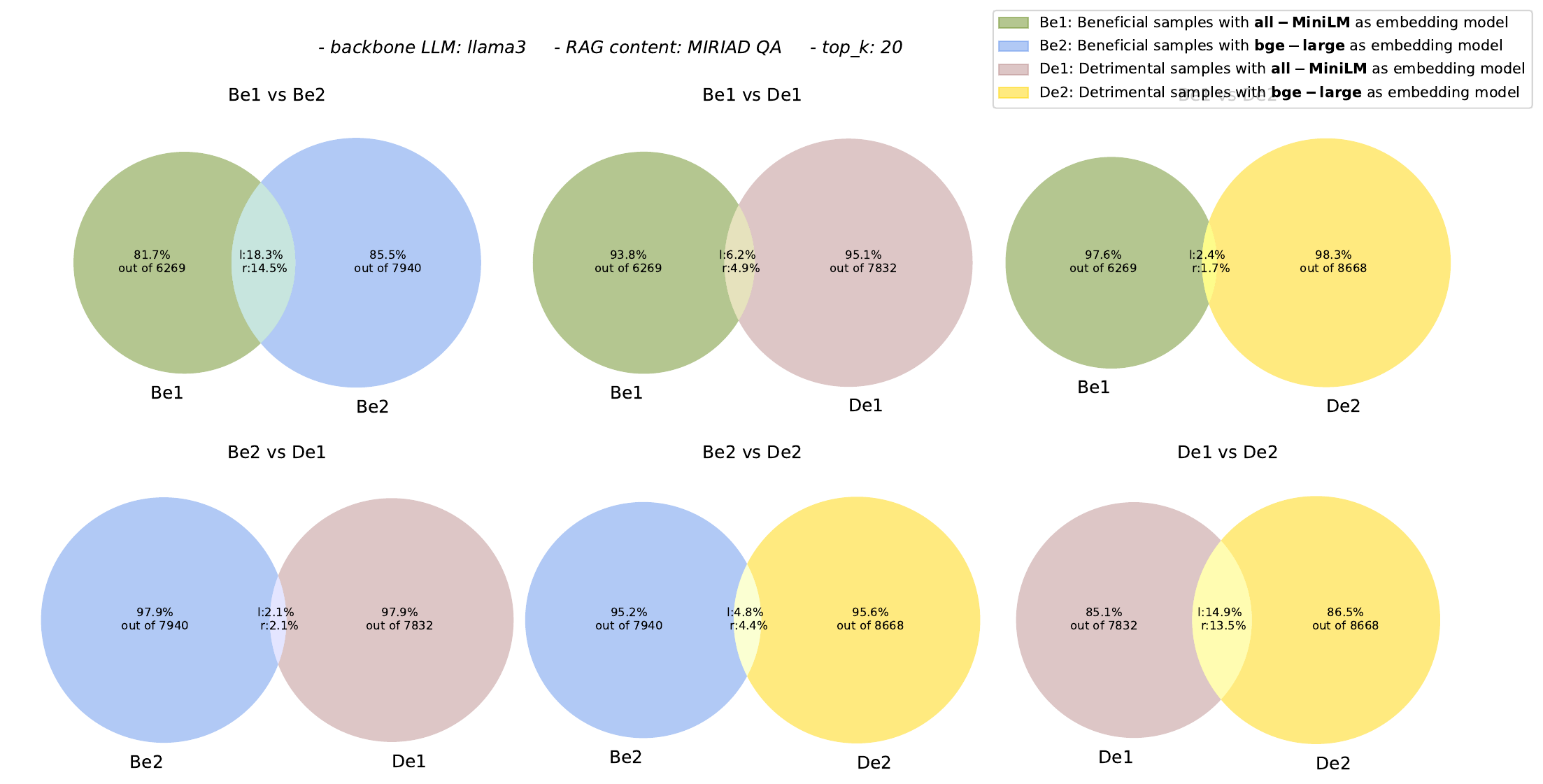}
    \caption{\textbf{Venn diagrams of beneficial and detrimental MIRAD samples for MedMCQA benchmark.} With fixed backbone LLM Llama-3.1-8B-instruct, top-k as 20, and different backbone embedding models.}
    \label{fig:llama3_qa_different_embed_top20}
\end{figure}

We further dissected RAG performance at the individual sample level through a controlled attribution analysis, where the isolated impact of individual retrieved MIRIAD QA pairs was examined. Each retrieved MIRIAD sample was classified based on its impact: helpful (facilitating a correct prediction that was otherwise incorrect), harmful (causing an incorrect prediction that would have otherwise been correct), or neutral (no discernible impact). Notably, the utility of individual retrieved samples was found to be highly dependent on the specific model configurations used. Venn diagram comparisons~(Fig.~\ref{fig:allminilm_qa_different_backbone_top3}--\ref{fig:llama3_qa_different_embed_top20}) revealed less than 20\% overlap in helpful or harmful retrievals across different embedding models or generative language models. This indicates that the beneficial or detrimental effect of individual retrieved samples is not intrinsic to the retrieved samples alone but is significantly context-dependent.



\section{Licensing}
\phantomsection 
In this paper, we use the Semantic Scholar Open Research Corpus (S2ORC) as the source of documents to generate our dataset. These documents are made available under the Open Data Commons Attribution License (ODC-By) v1.0 (https://opendatacommons.org/licenses/by/1-0/), which permits reuse and modification of the dataset, including for commercial use, provided that proper attribution is given. To construct our dataset, we used S2ORC documents as input to OpenAI's language models. The resulting model-generated outputs are owned by us, as per OpenAI’s Terms of Use, which also specify that outputs must not be used for medical diagnosis or decision-making about real individuals (https://openai.com/policies/terms-of-use/). Since our outputs are generated using both S2ORC documents and OpenAI's models, we release the dataset under the ODC-By v1.0 license, subject to the usage restrictions in OpenAI’s Terms of Use. \label{sec:licensing}

\section{Intended use} \label{sec:intendeduse}
At this stage, the outputs of this study and the provided assets are supplied exclusively for academic research and educational exploration. They have not been reviewed or cleared by any regulatory body, and accordingly must not be used for clinical decision-making or considered a certified medical device.

\end{document}